\documentclass{article} 
\usepackage[accepted]{icml2021}

\usepackage{amsmath,amsfonts,bm}

\def\eqref#1{equation~\ref{#1}}

\def\1{\bm{1}}

\DeclareMathAlphabet{\mathsfit}{\encodingdefault}{\sfdefault}{m}{sl}
\SetMathAlphabet{\mathsfit}{bold}{\encodingdefault}{\sfdefault}{bx}{n}

\DeclareMathOperator*{\argmax}{arg\,max}

\usepackage{hyperref}
\usepackage{url}
\usepackage{graphicx}
\usepackage{subcaption}
\usepackage{bbold}
\usepackage{adjustbox}
\usepackage{gensymb}
\usepackage{amsmath,amssymb,bm}
\usepackage[normalem]{ulem}
\usepackage{enumitem}

\usepackage{siunitx}
\usepackage{hyperref}
\usepackage{url}

\newlength{\vslen}
\usepackage{microtype}
\usepackage{graphicx}
\usepackage{booktabs} 

\usepackage[page]{appendix}

\icmltitlerunning{Fold2Seq for Protein Design}

\begin{document}

\twocolumn[
\icmltitle{Fold2Seq: A Joint Sequence(1D)-Fold(3D) Embedding-based Generative Model for  Protein Design}



\icmlsetsymbol{equal}{*}

\begin{icmlauthorlist}
\icmlauthor{Yue Cao}{goo,to,equal}
\icmlauthor{Payel Das}{goo}
\icmlauthor{Vijil Chenthamarakshan}{goo}
\icmlauthor{Pin-Yu Chen}{goo}
\icmlauthor{Igor Melnyk}{goo}
\icmlauthor{Yang Shen}{to}
\end{icmlauthorlist}

\icmlaffiliation{to}{Texas A\&M University}
\icmlaffiliation{goo}{IBM Research}

\icmlcorrespondingauthor{Y. Cao}{cyppsp@tamu.edu}
\icmlcorrespondingauthor{P.  Das}{daspa@us.ibm.com}
\icmlcorrespondingauthor{V. Chenthamarakshan} {ecvijil@us.ibm.com}
\icmlcorrespondingauthor{P.-Y. Chen}{pin-yu.chen@ibm.com}
\icmlcorrespondingauthor{I. Melnyk}{igor.melnyk@ibm.com}
\icmlcorrespondingauthor{Y. Shen}{yshen@tamu.edu}
\icmlkeywords{Protein Design, Generative Model}

\vskip 0.3in
]


\printAffiliationsAndNotice{$^*$Work primarily done during Yue Cao's internship at IBM Research.}



\begin{abstract}

{
 Designing novel protein sequences for a desired 3D topological fold is a fundamental yet  non-trivial task in protein engineering.
Challenges exist due to the complex sequence--fold relationship, as well as the difficulties to capture the diversity of the sequences (therefore structures and functions) within a fold. 
To overcome these challenges, we propose \textbf{Fold2Seq}, a novel transformer-based generative framework for designing protein sequences conditioned on a specific target fold. To model the complex sequence--structure relationship, Fold2Seq jointly learns a sequence embedding using a transformer and a fold embedding from the density of secondary structural elements in 3D voxels. On test sets with single, high-resolution and complete structure inputs for individual folds, our experiments  demonstrate improved or comparable performance of Fold2Seq in terms of speed, coverage, and reliability for sequence design, when compared  to existing state-of-the-art methods that include  data-driven deep generative models and physics-based RosettaDesign. The unique advantages of fold-based Fold2Seq,  in comparison to a  structure-based deep model and RosettaDesign, become more evident on three additional real-world challenges originating from low-quality, incomplete, or ambiguous input structures. Source code and data are available at \url{https://github.com/IBM/fold2seq}.

}
\end{abstract}

\section{Introduction}

Computationally designing protein sequences that fold into desired 3D structures has a broad range of applications ranging from therapeutics to materials \citep{kraemer2001computational}. 
Despite significant advancements in methodologies and computing power, this task known as inverse protein design still remains challenging, primarily due to the vast size of the sequence space as well as the difficulty of learning a function that maps from the 3D structure space to the 1D sequence space.

While majority of  data-driven approaches \citep{chen2019improve,o2018spin2,wang2018computational, ingraham2019generative, strokach2020fast} are focusing on designing sequences for a desired backbone structure, only a few works \citep{greener2018design, karimi2020novo} have studied protein design for a desired fold. A protein fold  is  defined by the spatial arrangement (or topology) of its 3D form of local segments called secondary structure elements or SSEs
\citep{hou2003global}. 
As protein structure is inherently hierarchical, a complete native structure can have fold combinations 
and a fold can be present in many protein structures.  
A simple fold or topological architecture can be highly adaptable, as shown by the low-sequence homology among its members, and the different functions they carry out \citep{basanta2020enumerative,chandra2001structural,boutemy2011structures}.  Therefore, a primary goal of \textit{de novo} protein design is  to generate a larger and more diverse set of protein structures than currently available yet still  consistent with a specific fold, which has proven to be a means for achieving new functions through design \citep{basanta2020enumerative, woolfson2015novo}. In  contrast, targeting a backbone structure \textit{per se} is known to restrict the diversity and novelty of the designs, as ``high-resolution protein backbone coordinates contain some memory of the original native sequence''~\citep{kuhlman2003design}.
Accordingly, an ensemble of structures is a better representative of a fold than a single structure,
 as it additionally captures the structural and therefore functional diversity within the fold. 
Compared to structure-based protein design, fold-based protein design carries additional challenges: the difficulties of learning a good fold representation for accurately capturing the diversity of the fold space and the complex fold-sequence relationship. 
 Despite the impressive  progress made by recent data-driven methods,  aforementioned challenges are not fully solved. \textbf{First}, the current fold representation methods are either hand-designed, or constrained  and do not capture the complete original fold space \citep{greener2018design,karimi2020novo,koga2012principles}, resulting in poor generalization or efficiency. 
 \textbf{Second}, the (1D) sequence encoding and the (3D) fold encoding are learned separately in previous methods, which makes two latent domains heterogeneous.  
Such  heterogeneity  across two domains actually increases the difficulty of learning the complex sequence--fold relationship. 
 To fill the aforementioned gaps, the \textbf{ main contributions} of this work are as follows:

1. We propose a novel fold representation, through first representing the 3D structure by the voxels of the SSE density, and then learning the fold representation through a transformer-based fold encoder. Compared to previous fold representations, this one has several advantages: it preserves the entire spatial information of SSEs in a scale-free manner,   does not need any pre-defined rules, 
and can be automatically extracted from a given protein structure. The density model also loosens the structure rigidity so that the structural variation and missing information is better handled. 

2. We employ a novel joint sequence--fold embedding learning framework into the transformer-based auto-encoder model. By learning a joint latent space between sequences and folds,  \textbf{Fold2Seq}, mitigates the heterogeneity between two different domains and is able to better capture the sequence-fold relationship, as reflected in the results.

3. We develop several novel fold-level assessment metrics. Using those, we  demonstrate that Fold2Seq has superior or comparable performance in perplexity,  sequence recovery rate, and  structure recovery rate, when compared to  competing methods including the state-of-the-art RosettaDesign and other neural-net models on the benchmark test set. More importantly, Fold2Seq-generated sequences  provide better coverage (diversity) within a specified fold.  Ablation study shows that this improved performance can be directly attributed to our algorithmic innovations. 

4. Experiments on real-world challenges   comprised of  low-resolution structures, structures with missing residues, and Nuclear Magnetic Resonance (NMR) ensembles further demonstrate the unique practical utility and versatility of Fold2Seq compared to the  structure-based  baselines.

\section{Related Work}

\textbf{Data-driven Protein Design.}
A significant surge of protein design studies that deeply exploit the data through modern artificial intelligence algorithms has been witnessed in the last three years. There appear a gallery of methods that focus on design protein sequences conditioned on the \textit{backbone structure} \citep{chen2019improve,o2018spin2,wang2018computational}. Recently, \citet{strokach2020fast}  formulated the inverse protein design as a constraint satisfaction problem
(CSP) and applied the graph neural networks for generating protein sequences conditioned on the residue-residue  distance map. \citet{ingraham2019generative} developed a graph-based transformer for generating protein sequences conditioned on the either rigid or flexible protein backbone information. Nevertheless, there are only a few studies that  investigated protein design  conditioned directly on the \textit{protein fold}.  
\citet{greener2018design} used the conditional variational autoencoder for generating protein sequences conditioned on a given fold. \citet{karimi2020novo}  developed a guided conditional Wasserstein  Generative Adversarial Networks (gcWGAN) also for the fold-based design.

\begin{figure*}[!htb]
    \centering
        \includegraphics[width=1.0\textwidth]{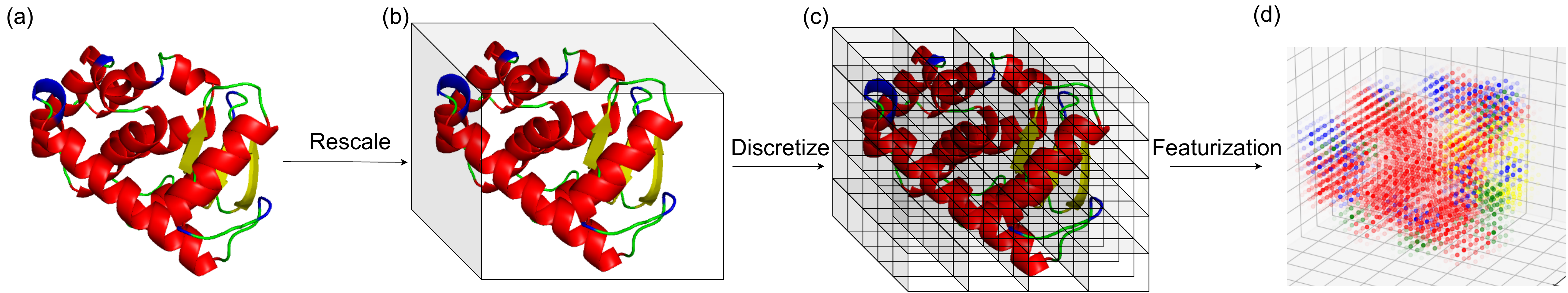}
        \vspace{-6mm}
    \caption{(a) The structure of T4 lysozyme (PDB ID 107L). The secondary structures are colored as: helices in red, beta sheets in yellow, loops in green and bend/turn in blue. (b) The structure is rescaled to fit the $40 \si{\angstrom} \times40 \si{\angstrom} \times40 \si{\angstrom}$ cubic box. (c)  The box is discretized into voxels. (d) Features of each voxel are obtained from the structure content of the voxel.  }
    \label{fig:pdb sturcture}
    \vspace{-2mm}
\end{figure*}

 \textbf{Protein Fold Representation.}
For an extensive overview of molecular representations, including those of proteins, please see  \citet{david2020molecular}. 
\citet{murzin1995scop} and \citet{orengo1997cath} manually classified protein structures in a hierarchical manner based on their structural similarity, resulting into 
one-hot encoding of the fold representations.
\citet{taylor2002periodic} represents a protein fold using a ``periodic table'' that  was later used for inverse fold design \citep{greener2018design}. However, it  considers three pre-defined folds ({$\alpha\beta\alpha$ layer, $\alpha\beta\beta\alpha$ layer and $\alpha\beta$ barrel}) for defining a fold space, limiting the spatial information content of the fold significantly. \citet{hou2003global} chose hundreds of representative proteins and calculated the similarity scores among them. This similarity matrix was then  converted into a distance matrix for kernel Principal Component Analysis (kPCA). A similar idea was used in \citet{karimi2020novo}  for inverse protein design.  This representation  needs a  set {(all-$\alpha$, all-$\beta$, $\alpha$/$\beta$ and $\alpha$+$\beta$)} of structures along with a similarity metric for defining a fold space, which may lead to biased or restricted representation of the fold space. Further, use of a similarity (or distance) matrix between fold pairs to learn fold representation, in principle,  may not preserve the detailed spatial information of the fold.
Finally, \citet{koga2012principles} summarized three rules that describe the junctions between adjacent secondary structure elements for a specific fold.  These rules are hand designed for a subset of structures, which makes the representation restricted to a small part of the fold space and offers limited generalizability during conditional sequence generation.

\textbf{Joint Embedding Learning.}
Joint embedding learning across different data modalities was first proposed by \citet{ngiam2011multimodal} on audio and video signals. Since then, such approaches have been widely used in cross modal retrieval or captioning  \citep{arandjelovic2018objects,gu2018look,peng2019cm,chen2018text2shape,wang2013learning,dognin2019adversarial}. In few/zero-shot learning, joint feature-label embedding was used 
\citep{zhang2016zero,socher2013zero}. Several studies have shown usefulness of learning joint embedding  for single modal classification 
\citep{ngiam2011multimodal,wang2018joint, toutanova2015representing}. Moreover, \citet{chen2018text2shape} used joint embedding learning for text to shape generation. Joint sequence--label embedding is also explored for or applied to molecular prediction/generation  \citep{10.1093/bioinformatics/btab198,das2018pepcvae}.

\section{Methods}
\vspace{\vslen}
\subsection{Background}
\vspace{\vslen}
A protein consists of a linear chain of amino acids (residues) that defines its 1D sequence. Chemical composition and interactions with neighboring residues  drive  the folding of a sequence into different secondary structure elements or SSEs (helix, beta-sheet, loop, etc., see Fig.~\ref{fig:pdb sturcture}(a)), that eventually forms a complete native 3D structure. %
A \textbf{protein fold} captures the structural consensus of the 3D topology and the composition of those secondary structure elements. 

\subsection{Fold Representation through 3D voxels of the SSE density}

\newcommand{\ax}[1]{\si{\angstrom}}

In \textit{de novo} protein design that we target, no backbone structure is assumed.  Instead, a topological ``blueprint'' (consistent with the desired fold) is given. And  initial backbone structures can be generated accordingly using fragment assemblies \citep{huang2016coming}. In this study we focus on generating fold representations once the structures are available and additionally explore the challenges from such ``blueprint'' input structures through three real-world challenges.

We hereby describe how we represent the 3D structure to explicitly capture the fold information, as illustrated in Fig.~\ref{fig:pdb sturcture}.  The position (3D coordinates) of each residue is represented  by its $\alpha$-carbon.  For a given protein of length $N$, we first translate the structure 
to match its  center of geometry ($\alpha$-carbon) with the origin of the coordinate system. We then rotate the protein around the origin to let the first residue  be on the negative side of $z$-axis (principal component-based orienting was also explored as in \textbf{Training and Decoding Strategy}).  We denote the resulting residue coordinates as $\bm{c}_1, \bm{c}_2, ... , \bm{c}_N$. The secondary structure label to each residue is assigned based on their SSE assignment \citep{kabsch1983dictionary} in Protein Data Bank \citep{berman2000protein}. 
We consider 4 types of secondary structure labels: helix, beta strand, loop and bend/turn.  In order to consider the distribution of different secondary structure labels in the 3D space, we discretize the 3D space into voxels. Due to the scale-free definition of a protein fold,
we  rescale the original structure, so that it fits into a fixed-size cubic box. Based on the distribution of  sizes of single-chain, single-domain proteins in the CATH database \citep{sillitoe2019cath}, we choose a $40 \si{\angstrom} \times40 \si{\angstrom} \times40 \si{\angstrom}$ box with each voxel of size $2 \si{\angstrom} \times 2 \si{\angstrom} \times 2\si{\angstrom}$. We denote the scaling ratio as $\bm{r} \in \mathcal{R}^3$. 
For voxel $i$,  we denote the coordinates of its center as $\bm{v}_i$. We assume that the contribution of residues $j$ to  voxel $i$ follows a Gaussian form:
\begin{equation}\label{eq:fold_features}
    \bm{y}_{ij} = \exp (-\frac{||\bm{c}_j \odot \bm{r} - \bm{v}_i||_2^2}{\sigma^2}) \cdot \bm{t}_j,
\end{equation}
where $\bm{t}_j \in \{0,1\}^4$ is the one-hot encoding of the secondary structure label of amino acid $j$. The standard deviation is chosen to be $2\si{\angstrom}$. We sum up all residues together to obtain the final  features of the voxel $i$: $\bm{y}_i = \sum_{j=1}^N \bm{y}_{ij}$.
The fold representation $\bm{y} \in \mathcal{R}^{20 \times 20 \times 20 \times 4}$  is the 4D tensor of $\bm{y}_i$ over all $20\times 20
\times 20$ voxels. 
This fold representation using 3D SSE densities better captures scale-free SSE topologies that define folds, while removing fold-irrelevant structure details. It results in sequence generation that explores the  sequence space available to a specific fold more widely (as shown in experiments).

\subsection{Fold2Seq with Joint Sequence--Fold Embedding}

\begin{figure*}[!htb]
    \centering
     \includegraphics[width=0.93\textwidth]{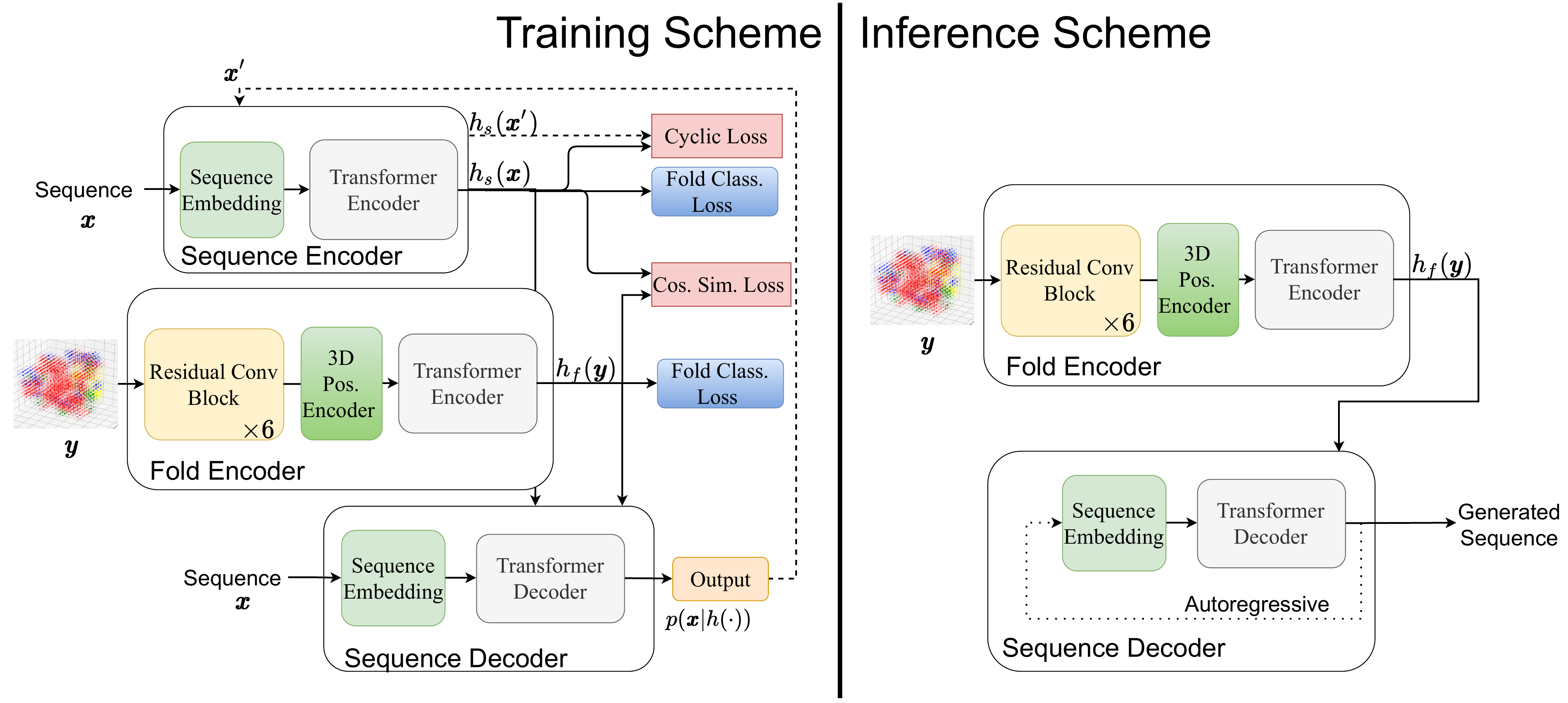}
        \vspace{-4mm}
    \caption{The architecture of the Fold2Seq model during the training and inference stages. (\textbf{Training Scheme}): During training, the model includes three major components: (top) Sequence Encoder, (middle) Fold Encoder and (bottom) Sequence Decoder.  The dashed arrows represent the process for getting cyclic loss. (\textbf{Inference Scheme}): During the inference, the model  only needs the fold encoder and the sequence decoder for conditionally decoding sequences.} 
    \label{fig:architecture}
    \vspace{-2mm}
\end{figure*}

\paragraph{Model Architecture.} 
In the training stage, our model consists of three major components: a sequence encoder: $h_s(\cdot)$, a fold encoder: $h_f(\cdot)$ and a sequence decoder: $p(\bm{x}|h(\cdot))$, as shown in 
Fig.~\ref{fig:architecture} (Left). 

(i) \textbf{Sequence Encoder/Decoder.} Both sequence encoder and decoder are implemented using the vanilla transformer model and a vanilla sequence embedding module (learnable lookup table + sinusoidal positional encoding), as described in   \citet{vaswani2017attention}. All training sequences are padded to the maximum length $N_s$ of 200, as  77\% of single-domain sequence lengths in the CATH dataset are $\leqslant 200$.

(ii) \textbf{Fold Encoder.} A fold representation $\bm{y} \in \mathcal{R}^{20 \times 20 \times 20 \times 4}$ will go through a fold encoder, which contains 6 residual blocks followed by a 3D positional encoding. Each residual block has two  3D-convolutional layers ($3\times3\times3$) and batch normalization layers. The 6 residual blocks transform $\bm{y}$ to a tensor with the shape $5 \times 5 \times 5 \times d$, where $d$ is  the hidden dimension. The 3D positional encoding is a simple 3D extension of the  sinusoidal encoding described in the vanilla transformer model, 
as shown in Section \ref{append: structure} of the appendix. After the positional encoding, the 4D tensor is flattened to be 2D with the shape $125 \times d$, as the input of a transformer encoder. The output of the transformer encoder, $h_f(\bm{y})$, is the latent fold representation of $\bm{y}$.

We propose a simple fold-to-sequence reconstruction loss based on the auto-encoder model: $\textbf{RE}_f=p(\bm{x}| h_f(\bm{y}))$. However, as mentioned earlier, 
training based on  $\textbf{RE}_f$ alone suffers  from  
the  heterogeneity of $\bm{x}$ and $\bm{y}$. 
To overcome this challenge, we first encode the sequence $\bm{x}$ through the sequence encoder into the latent space as $h_s(\bm{x})$, which could be done through a simple sequence-to-sequence reconstruction loss: $\textbf{RE}_s=p(\bm{x}| h_s(\bm{x}))$.  We then learn a joint latent space  between $h_f(\bm{y})$ and $h_s(\bm{x})$ through a novel sequence-fold embedding learning framework with additional losses detailed below.

\textbf{Joint Embedding Learning.}
Typically, learning a joint embedding across two domains needs two intra-domain losses and one cross-domain loss \citep{chen2018text2shape}. An intra-domain loss forces two semantically similar samples from the same domain to be close to each other in the latent space, while a cross-domain loss forces two semantically similar samples in different domains to be closer.

In our case, the meaning of `semantically similar' is that the proteins should have the same fold(s). Therefore, we consider a supervised learning task for learning intra-domain similarity: fold classification. 
Specifically, the outputs of both encoders: $h_f(\bm{y}) \in \mathcal{R}^{l_f \times d}$ and $h_s(\bm{x}) \in \mathcal{R}^{l_s \times d}$ will be averaged along $l_f$ and $l_s$ dimensions, followed by a MLP+softmax layer to perform fold classification (shown as two blue blocks in Fig.~\ref{fig:architecture}), where  $l_s$ and $l_f$  are the length of the sequence and the fold, respectively. The parameters of the two MLP layers are shared. The  category labels follow the fold (topology) level of hierarchical protein structure classification in CATH4.2 dataset \citep{sillitoe2019cath} (see Section~\ref{sec:datasets}). As a result, we propose the following two intra-domain losses: $\textbf{FC}_f$ and $\textbf{FC}_s$, i.e. the cross entropy losses of fold classification from   $h_f(\bm{y})$ and $h_s(\bm{x})$ respectively.
The benefits of these two classification tasks are as follows: First, it will force the fold encoder to learn the fold representation. Second, as we perform the same supervised learning task on the latent vectors from two domains, it will not only learn the intra-domain similarity, but also cross-domain similarity. In contrast, without explicit cross-domain learning, the two latent vectors $h_f(\bm{y})$, $h_s(\bm{x})$ could still have minimal alignment between them. 

In the transformer decoder, each element in the \textbf{non}-self attention matrix is calculated by the cosine similarity between the latent vectors from the encoder and the decoder, respectively. Inspired by this observation, we  maximize the cosine similarity (shown as the `Cosine Similarity' in Fig.~\ref{fig:architecture}) between $h_f(\bm{y}) \in \mathcal{R}^{l_f \times d}$ and $h_s(\bm{x}) \in \mathcal{R}^{l_s \times d}$ as the cross-domain loss.  We first calculate the matrix-product  between $h_f(\bm{y})$ and $h_s(\bm{x})$ as  $\bm{Q} = h_f(\bm{y}) \cdot h_s(\bm{x})^T, \bm{Q} \in \mathcal{R}^{l_f \times l_s}$. The $i$th row in $\bm{Q}$ represents the similarity between  $i$th position in the fold and every position of the sequence.
We would like to find the best-matching sequence piece with each position in the fold. To achieve this, the similarity matrix $\bm{Q}$ first goes through a row-wise average pooling with kernel size $k$, followed by the row-wise $\max$ operation:
\begin{equation}
    \bm{q} = \max_{row}(\text{AvgPool}^k_{row}(\bm{Q})), \bm{q} \in \mathcal{R}^{l_f \times 1},
\end{equation}
where $row$ indicates row-wise operation. We choose $k=3$, \textit{i.e.} the scores of every 3 contiguous positions in the sequence will be averaged. We finally average over all positions in the fold to get the final similarity  score: $\textbf{CS} = \text{mean} (\bm{q})$.

Besides the cosine similarity loss, inspired by the earlier CycleGAN work \citep{zhu2017unpaired}, we add a cyclic loss (shown as  the  red block of ``Cyclic Loss'' in Fig.~\ref{fig:architecture}.) to be another term of our cross-domain loss.  Specifically, we take the argmax of the output of fold-to-sequence model: $\bm{x}^\prime = \argmax p(\bm{x}|h_f(\bm{y}))$, and send it  back to the sequence encoder  for generating the cyclic-seq latent state:  $h_s(\bm{x}^\prime)$ (shown as the dashed line in Fig.~\ref{fig:architecture}).  This cyclic-seq latent state  will compare with the native seq latent state $h_s(\bm{x})$ through the square of the L2 distance:
\begin{equation}
    \textbf{CY} = ||h_s(\bm{x}^\prime)-h_s(\bm{x})||^2_2 
\end{equation}
To summarize, the complete loss  objective is the following:
    \begin{equation}
    L = \lambda_1 \textbf{RE}_f + \lambda_2 \textbf{RE}_s + \lambda_3 \textbf{FC}_f + \lambda_4 \textbf{FC}_s + \lambda_5 (\textbf{CY} - \textbf{CS}),
\end{equation}
where   $\lambda_1 $ through $\lambda_5$ are the hyperparameters for controlling the relative importance among these losses.

\textbf{Training and Decoding Strategy.}
During experiments we found that, if the sequence encoder and the fold encoder were trained together, the fold encoder had little parameter improvement while the sequence encoder dominated the training. To overcome this issue,   
we consider a two-stage training strategy. In the first stage, we train the sequence-to-sequence model regularized by the sequence intra-domain loss: $L_1 =  \lambda_2 \textbf{RE}_s +  \lambda_4 \textbf{FC}_s.$  
After the first stage is finished, we start the second training stage. We train the fold-to-sequence model regularized by the fold intra-domain loss and the cross-domain loss while keeping the sequence encoder frozen: $L_2 = \lambda_1 \textbf{RE}_f + \lambda_3 \textbf{FC}_f + \lambda_5 (\textbf{CY} - \textbf{CS} ).$
The comparison between the one-stage training and two-stage training strategies are described in details in Section \ref{append: loss} in the appendix.
 
We  implement our model in Pytorch \citep{paszke2019pytorch}. Each transformer block has 4 layers and $d=256$ latent dimensions. In order to increase the robustness of our model for rotated structures, we augment our training data by right-hand rotating the each structure by $90^\circ$, $180^\circ$ and $270^\circ$ along each axis (x,y,z). As a result, we augment our training data by $3\times3=9$ folds. While orienting proteins along principal axes is better using global shapes, we find neither orientations along principal axes nor denser augmentations (45\degree)  empirically  boosted the model performance (See Result in Sec. 4). The learning rate schedule follows  the original transformer paper \citep{vaswani2017attention}. We use the exponential decay \citep{blundell2015weight} for $\lambda_5=1/2^{\text{\#epoch}-e}$ in the loss function, while  $\lambda_1$ through $\lambda_4$ and $e$ are tuned based on the validation set, resulting in  
$\lambda_1=1.0,\lambda_2=1.0,  \lambda_3=0.02, \lambda_4=1.0, e=3$. We train our model on 2 Tesla K80 GPUs, with batch size 128. In every training stage we train up to 200 epochs with an early stopping strategy based on the validation loss. 

During inference, one only needs the fold encoder and the sequence decoder for conditional sequence generation 
(Fig.~\ref{fig:architecture} (Right)). Top-$k$ sampling strategy \citep{fan2018hierarchical} is used for sequence generation, where $k$ is tuned to be 5  based on the validation set.

\subsection{Benchmark Datasets}\label{sec:datasets}
\vspace{\vslen}

We used protein structure data from CATH 4.2 \citep{sillitoe2019cath} filtered by 100\% sequence identity. 
We remove proteins that (1) are multi-chain or  non-contiguous in sequence; (2) contain other than 20 natural amino acids; or (3) have length longer than 200.  We randomly split the dataset  at the fold level into 95\%, 2.5\%, 2.5\% as dataset (a), (b) and (c), respectively, which means that the three datasets have non-overlapping folds.  We further randomly split the dataset (a) at the structure level into 95\%, 2.5\% and 2.5\% as dataset (a1), (a2) and (a3), respectively. Datasets (a1),  (a2), and (a3) have overlapping folds. We use dataset (a1) as the training set, (b)+(a2) as the validation set, (a3) as the In-Distribution (ID) test set and (c) as the Out-of-Distribution (OD) test set. The folds of the ID test set overlap with the training set, whereas the folds of the OD test set do not. 
Statistics of these datasets are presented in Section \ref{append:datasets} in the appendix.

To quantitatively measure their difficulty levels, we calculate the  \textbf{a}veraged \textbf{m}aximum \textbf{s}equence  \textbf{i}dentity (\textbf{amsi}) between a given test set $\bm{T}$ and the training set as: $\text{amsi}_{\bm{T}} = \frac{1}{|D_{\bm{T}}|}\sum_{j \in D_{\bm{T}}} \max_{k \in D_{\text{train}}}(\text{SIM}(\bm{x}_j, \bm{x}_k))$, where $D_{\text{train}}$ and $D_{\bm{T}}$ are the training and test ($\bm{T}$) set, respectively; $\text{SIM}(\bm{x}_j, \bm{x}_k)$ 
is the sequence identity (See Section \ref{app:seq_id} in the appendix) between sequence $\bm{x}_j$ and $\bm{x}_k$. We found  $ \text{amsi}_{\text{ID}}=36.3\%$ and $ \text{amsi}_{\text{OD}}=16.3\%$, showing that 
the OD test set represents a much more difficult generalization task compared to the ID. 

\vspace{-0.5em}
\subsection{Assessment Metrics}
Ideally, the most appropriate and rigorous criteria for evaluating fold-based protein design methods is to check the consistency between  the structures of the generated sequences and  the desired fold. However, as protein structure prediction is very computationally expensive, and similar sequences usually indicate similar folds or structures, many earlier structure-based methods \citep{ingraham2019generative, madani2020progen} report performance in the sequence domain. Here, considering that a fold is comprised of multiple structures, we define four fold-level metrics that is able to assess the quality of the designed sequences for a desired fold. 
For a test set $D_{\textbf{T}}$, we use $i \in D_{\textbf{T}}$ to represent   a fold in $D_{\textbf{T}}$ and $j \in D_{\textbf{T}}$ (or $k \in D_{\textbf{T}}$) to represent a protein in $D_{\textbf{T}}$.
\vspace{-1.0em}
\paragraph{(Fold-level) Per-residue Perplexity.} Based on \citep{ingraham2019generative, madani2020progen},
the structure-level per-residue PerPLexity ($ppl$) for test $D_{\textbf{T}}$ is defined as: $ppl_{\text{structure}}(D_{\textbf{T}}) = \exp  (-\frac{1}{|D_{\textbf{T}}|}\sum_{j \in D_{\textbf{T}}} \frac{1}{L_j}\log p(\bm{x}_j|\bm{y}_j)$), where  $L_j$ is the length of sequence $j$. Here we consider  per-residue perplexity for fold $i$: 
\begin{equation}
   ppl_{\text{fold}}(i) = \exp  (-
   \frac{1}{|\mathcal{S}_i|}\sum_{j \in \mathcal{S}_i}
   \max_{k \in \mathcal{S}_i}( \frac{1}{L_k}\log p(\bm{x}_k|\bm{y}_j)),  
\end{equation}
where $\mathcal{S}_i$ is set of structures in fold $i$. We compute  the mean and  standard deviation of $ppl_{\text{fold}}(i)$ over all folds in $D_{\textbf{T}}$.
\vspace{-1.0em}
\paragraph{(Fold-level) Sequence Recovery.}
We define the set of the generated sequences (structures) conditioned on structure $j$ as $\mathcal{G}_j$. 
In structure-based design, we usually define the Sequence Recovery rate ($sr$) for $\bm{y}_j$ as $sr_{\text{structure}}(j) = \frac{1}{|\mathcal{G}_j|}\sum_{g \in \mathcal{G}_j} \text{SIM}(\bm{x}_g, \bm{x}_j) $. Here we consider the fold-level sequence recovery rate for fold $i$:
\begin{equation}
    sr_{\text{fold}}(i) = \frac{1}{|\mathcal{S}_i|} \sum_{j \in \mathcal{S}_i} \frac{1}{|\mathcal{G}_j|}\sum_{g \in \mathcal{G}_j} \max_{k \in  \mathcal{S}_i} \{\text{SIM}(\bm{x}_g, \bm{x}_k)\}.
\end{equation}
\vspace{-1.0em}
\paragraph{(Fold-level) Coverage (Diversity).} We also measure how the generated sequences from a single (or few) representative structure(s) could capture the diversity of sequences (thus of structures and functions) within a fold. 
To do so,  for fold $i$, we randomly pick \textbf{one} structure $k$ from $\mathcal{S}_i$ as the representative. We then measure how many sequences within that fold are captured by the generated sequences conditioned on the representative. As a result, we define the COVerage ($cov$) for fold $i$ as:
\begin{equation} 
        cov_{\text{fold}}(i) =
    \frac{1}{|\mathcal{S}_i|}\sum_{j \in \mathcal{S}_i}\mathbb{1}( \max_{g \in \mathcal{G}_k} \{\text{SIM}(\bm{x}_g, \bm{x}_j)\} \geqslant 30\%).
\label{eq:cov}
\end{equation}
We use the rule of thumb: two sequences 
likely belong to the same fold if their identity is above 30\% \citep{rost1999twilight}. 

\vspace{-1.0em}
\paragraph{(Fold-level) Structure Recovery.} We last  assess design accuracy directly in the structure domain. In structure-based design, similar to the sequence recovery, the sTructure Recovery ($tr$) rate is defined as: $tr_{\text{structure}}(j) = \frac{1}{|\mathcal{G}_j|}\sum_{g \in \mathcal{G}_j} \text{TM}(\bm{y}_g, \bm{y}_j) $, where $\text{TM}(\bm{y}_g, \bm{y}_k)$ is the TM-score \citep{zhang2004scoring} 
between structures $\bm{y}_g$ and $\bm{y}_k$. Here we extend it to fold-level structure recovery:
\begin{equation}
    tr_\text{fold}(i) = \frac{1}{|\mathcal{S}_i|} \sum_{j \in \mathcal{S}_i} \frac{1}{|\mathcal{G}_j|}\sum_{g \in \mathcal{G}_j} \max_{k \in  \mathcal{S}_i} \{\text{TM}(\bm{y}_g, \bm{y}_k)\}.
\end{equation}

We used the iTasser Suite \citep{yang2015tasser}, one of the state-of-the-art protein structure prediction software, to predict the structure of the designed sequences. For all metrics in the sequence domain, we have set $|\mathcal{G}_j|=100$ for every $j \in D_{\textbf{T}}$. However, as iTasser usually takes at least one day for predicting the structure of a single protein, for $tr_\text{fold}$ in the structure domain, we use  $|\mathcal{G}_j|=1$ for every $j \in D_{\textbf{T}}$.
We also include the performances of different methods based on the structure-level metrics in Section \ref{app:str_level} in the appendix. 

\subsection{Baseline Methods}

\textbf{Data-driven.} We  consider two data-driven \textit{fold-based}  methods that design sequences conditioned on a desired fold: cVAE \citep{greener2018design} and gcWGAN \citep{karimi2020novo}. We also consider a recent   \textit{structure-based} method, Graph\_trans \citep{ingraham2019generative}, that uses graph specification on the backbone structure as input and has shown to outperform earlier structure-based methods in terms of the structure-level metrics. We used Graph\_trans conditioned on the  flexible backbone for comparison. \textbf{Physics-based.} We then consider the state-of-the-art principle-driven method, RosettaDesign\footnote{RosettaDesign uses MCMC sampling and energy calculation to search for best sequences. The input to RosettaDesign consists of the backbone of the native structure and the SSE of each residue.} \citep{huang2011rosettaremodel}.

\section{Experiments on Benchmark Test Sets}
\vspace{\vslen}

\paragraph{Perplexity and Sequence Recovery Comparison.}
We first compare $ppl_{\text{fold}}$ of Fold2Seq with  those of the baseline methods 
(except RosettaDesign, as $ppl_{\text{fold}}$ is not applicable to it). 
For reference, we also show the per-residue perplexity under the uniform distribution and the frequencies through all natural sequences in UniRef50 \citep{suzek2015uniref}. We do not report standard deviation on these perplexities 
as they are unconditional distributions. Performances on two test sets are  summarized in Table \ref{tab:perplexity}, showing that Fold2Seq has the smallest $ppl_{\text{fold}}$ on ID test set and the second smallest on OD test set.  We have also tested different data augmentation strategies including orientations along principal axes and denser augmentations (45\degree). Neither strategy significantly boosted the model performance for the OD set ($ppl_\text{fold}$: 12.2 (2.7) and 12.0 (2.5), respectively). 

\begin{table}[!htb]
\caption{Performance of different methods assessed by (a) Avg. $ppl_\text{fold}$ (std. dev.) and (b)  Avg. $sr_{\text{fold}}$ (std. dev.) (\%).}
\begin{subtable}[h]{0.5\textwidth}
\centering
\setlength{\arrayrulewidth}{0.5mm}
\caption{}
\resizebox{0.55\textwidth}{!}{%
\begin{tabular}{r  c  c }
\hline
Model & ID Test & OD Test \\
\hline
Uniform & 20.0 & 20.0  \\
Natural & 18.0 & 18.0 \\ 
\hline
 cVAE & 13.2 (2.2) & 15.2 (2.3)  \\
gcWGAN & 12.3 (2.3)  & 14.3 (2.5) \\
Graph\_trans & 9.6 (2.9) & \textbf{11.5 (3.3)} \\
Fold2Seq &  \textbf{9.0 (5.3)} & 12.0 (2.4)  \\ \hline 
 \end{tabular}%
}
\label{tab:perplexity}
\end{subtable}
\begin{subtable}[h]{0.5\textwidth}
\centering
\caption{}
\setlength{\arrayrulewidth}{0.5mm}
\resizebox{0.76\textwidth}{!}{%
\begin{tabular}{r   c  c }
\hline
Model & ID Test & OD Test   \\
\hline
Random across two  folds  & 12.8 (7.9) & 12.8 (7.9) \\
\hline
cVAE & 18.2 (6.7) & 17.3 (5.2) \\
gcWGAN &  20.6 (5.4) & 19.2 (3.7) \\
RosettaDesign & 22.1 (5.7) &  22.3 (3.5) \\    
Graph\_trans & \textbf{28.8 (11.3)} & \textbf{27.1 (4.0)} \\
Fold2Seq &  27.2 (6.3)  &   25.2 (3.2)\\
\hline
Random within same fold & 39.1 (9.4)  & 39.1 (9.4) \\  \hline          
 \end{tabular}%
}
\label{tab:seq_recover_mean}
\end{subtable}
\end{table}

Next, we compare different methods for recovering the native sequences within  a desired fold.  Also, for comparison, we calculate the expected similarity between two random sequences in our whole dataset belonging to two different folds and belonging to the same folds. The results are summarized in Table \ref{tab:seq_recover_mean}. Overall, Graph\_trans 
and Fold2Seq outperform other methods by a big margin, while Graph\_trans shows slightly better performance than Fold2Seq. This is because  Graph\_trans is the only baseline that utilizes more high-resolution structural information beyond the fold level as the input. However, a structure-based method may not capture the similarity and diversity within the fold space.  To highlight this point, we 
 use t-SNE to visualize the fold embeddings $\bm{h}$ after the fold encoder for the proteins in the OD test sets. The results  in Section \ref{append: tsne} in the appendix evidently show  that the embeddings of same-fold proteins from Graph\_trans are less clustered than those from Fold2Seq. 

\vspace{-1em}
\paragraph{Coverage (Diversity).}
Coverage, as defined in Eq \ref{eq:cov}, is shown in Table \ref{tab:seq_coverage}.  
We split the folds within the testsets based on the number of sequences within each fold ($|\mathcal{S}_i|$) using a cutoff of 3.  
Overall, Fold2Seq shows better coverage, compared to other baselines.
In most cases, coverages on more diverse folds ($|\mathcal{S}_i|$ $>$ 3) have smaller standard deviations due to large $|\mathcal{S}_i|$ in the denominator of Eq. \ref{eq:cov}.
We then directly compare Fold2Seq with Graph\_trans by counting the number of folds for which  Fold2Seq yields better $cov_{\text{fold}}(i)$. As shown in Table \ref{tab:coverage_head2head}, Fold2Seq provides better coverage  in 68\%-88\% folds, implying that the proposed method can better capture the diversity within a fold compared to Graph\_trans. 

Moreover, we compare with an alternative version of Graph\_trans: Graph\_trans\_all, which 
conditions on each structure within a fold and then
combines the sequences generated over all conditions (instead of one) 
for calculating the coverage.  Though such an approach treats structure inputs separately and do not target what makes diverse structures common to a fold or distinguished across folds (evident in visualization of learned embeddings in Section \ref{append: tsne} in the appendix).  
Table \ref{tab:seq_coverage} shows that Fold2Seq outperforms Graph\_trans\_all in most cases except in the OD test set with $|\mathcal{S}_i|$ $>$ 3.

\begin{table}[!htb]
\caption{(a) Avg. $cov_{\text{fold}}$ (std. dev. in \%). (b) Fold2Seq(f) and Graph\_trans(g) head-to-head coverage comparison.}
\begin{subtable}{0.5\textwidth}
\centering
\caption{}
\begin{adjustbox}{width=0.95\textwidth}
\setlength{\arrayrulewidth}{0.5mm}
\begin{tabular}{r r r | r r}
\hline
& \multicolumn{2}{c}{ID Test} & \multicolumn{2}{c}{OD Test}   \\
\hline
Subset &   $|\mathcal{S}_i|$ $\leqslant $ 3 &  $|\mathcal{S}_i|$ $>$ 3  &   $|\mathcal{S}_i|$ $\leqslant $ 3 &  $|\mathcal{S}_i|$ $>$ 3\\
\hline
cVAE & 16.2 (17.3) &  13.3 (16.1) & 15.2 (16.3) & 11.3 (12.4)\\
gcWGAN & 18.9 (15.3)  & 20.5 (21.2) & 17.3 (13.4) & 15.3 (12.8) \\ 
Graph\_trans & 19.4 (28.9) & 24.1 (25.1) & 26.9 (32.5) & 20.2 (19.8)\\
$\text{Graph\_trans}\_{\text{all}}$ &  28.9 (32.3) & 25.3 (30.1) & 30.2 (25.2) & \textbf{21.3 (23.7)}  \\
RosettaDesign & 20.3 (17.3) &  17.3 (16.2) &  21.2 (20.3) & 17.5 (18.9) \\  
Fold2Seq & \textbf{32.9 (33.5)} & \textbf{28.9 (27.8)} & \textbf{34.3 (38.3)} & 20.7 (17.7)  \\
\hline
 \end{tabular}
 \end{adjustbox}
\label{tab:seq_coverage}
\end{subtable}
\begin{subtable}{0.5\textwidth}
\centering
\caption{}
\begin{adjustbox}{width=0.95\textwidth}
\setlength{\arrayrulewidth}{0.5mm}
\begin{tabular}{r r r | r r}
\hline
& \multicolumn{2}{c}{ID Test} & \multicolumn{2}{c}{OD Test}   \\
\hline
Subset &   $|\mathcal{S}_i|$ $\leqslant $ 3 &  $|\mathcal{S}_i|$ $>$ 3  &   $|\mathcal{S}_i|$ $\leqslant $ 3 &  $|\mathcal{S}_i|$ $>$ 3\\
\hline
\#$cov^\mathrm{f}_\mathrm{fold}(i)>cov^\mathrm{g}_\mathrm{fold}(i)$  & 104 & 53 & 13  & 8\\
Total  \#folds & 118 &  78 &18 & 10 \\
Ratio & 0.88 & 0.68 & 0.72 & 0.80 \\
\hline
 \end{tabular}
 \end{adjustbox}
\label{tab:coverage_head2head}
\end{subtable}
\end{table}
\vspace{-1em}

\vspace{\vslen}
\begin{figure*}[!hbt]
    \centering
    \begin{subfigure}[b]{0.31\textwidth}
     \includegraphics[width=\textwidth]{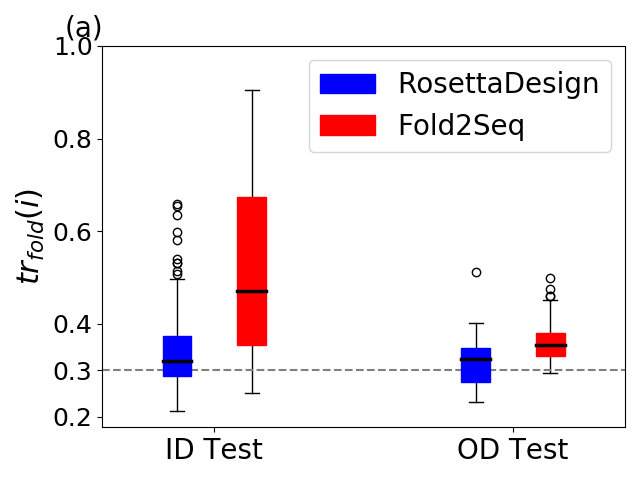}
     \end{subfigure}
    \begin{subfigure}[b]{0.31\textwidth}
    \includegraphics[width=\textwidth]{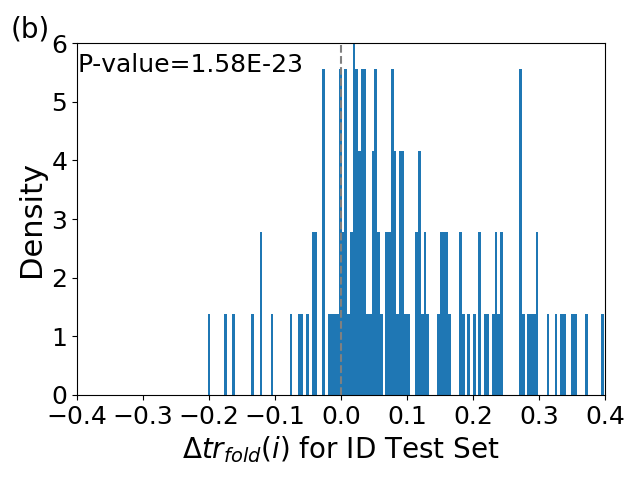}
    \end{subfigure}
        \begin{subfigure}[b]{0.31\textwidth}
    \includegraphics[width=\textwidth]{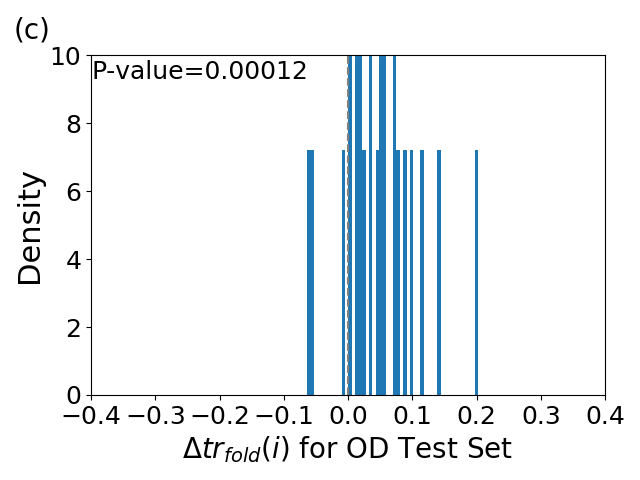}
    \end{subfigure}
        \begin{subfigure}[b]{0.31\textwidth}
    \includegraphics[width=\textwidth]{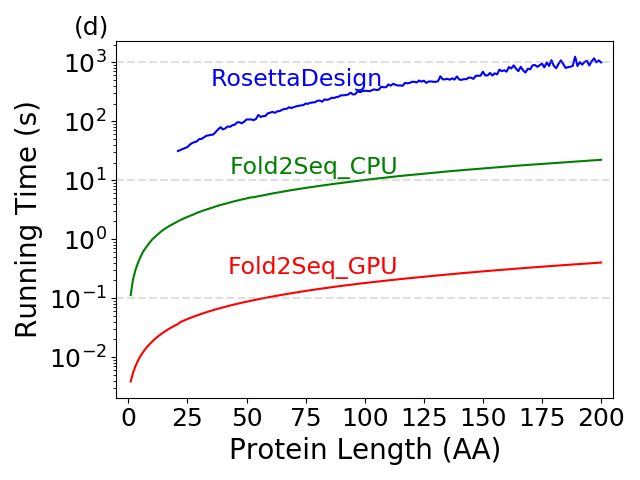}
    \end{subfigure}
        \begin{subfigure}[b]{0.31\textwidth}
    \includegraphics[width=\textwidth]{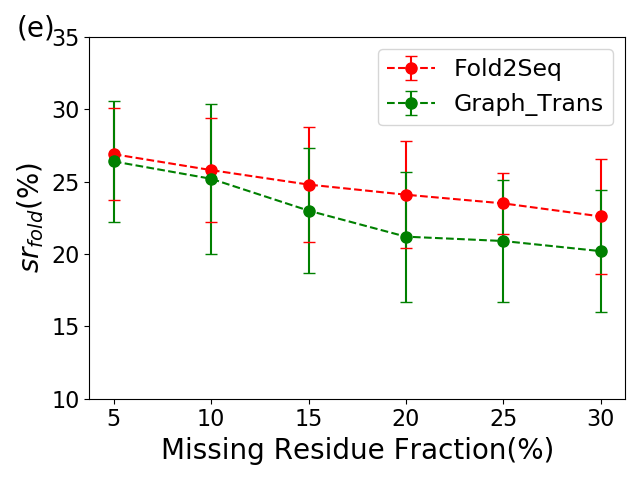}
    \end{subfigure}
    \vspace{-4mm}
          \caption{(a). $tr_{\text{fold}}(i)$ distributions  of RosettaDesign and Fold2Seq. (b, c). The distributions of $\Delta tr_{\text{fold}}(i)$ for ID test set and OD test set, respectively.  (d). Run time of Fold2Seq and RosettaDesign for generating one protein sequence: CPU: Intel Xeon E5-2680 v4 2.40GHz, GPU: Nvidia Tesla K80. (e). Avg. $sr_{\text{fold}}(i)$ for the OD test set  with a continuous stretch of  missing residues.}
            \label{fig:seq_gen}
\vspace{\vslen}            
\end{figure*}
\vspace{-0.5em}

\paragraph{Structural Recovery Comparison with RosettaDesign.}
Besides sequence-domain assessments, we examine if the  structure of a  Fold2Seq-generated sequence is of the same or similar folds to the native structure. Due to the associated computational expense, 
we limit structure predictions to proteins designed by Fold2Seq and RosettaDesign. 
We first compare the distributions of $ts_{\text{fold}}(i)$  on two test sets (ID and OD); results are shown in Fig.~\ref{fig:seq_gen}(a).  Fold2Seq shows significant improvement against RosettaDesign. The performance of Fold2Seq on ID test set is better than that on OD,
 thus matching their expected difficulty levels.  RosettaDesign performs similarly on both sets due to its physics-based nature that does not rely on learning  from a training set.

To quantitatively measure the performance difference between the two methods, we define $\Delta ts_{\text{fold}}(i) = ts^{\textbf{Fold2Seq}}_{\text{fold}}(i) - ts^{\textbf{Rosetta}}_{\text{fold}}(i)$ and  perform one-sided one-sampled t-test over $\Delta ts_{\text{fold}}$, with the null hypothesis as ``$\Delta ts_{\text{fold}} \leqslant  0.0$" on two test sets. The resulting $\textbf{P-value}_{\text{ID}}=1.58\text{E}-23$ and $\textbf{P-value}_{\text{OD}}=0.00012$ demonstrate that, overall,  Fold2Seq can  generate more reliable structures compared to RosettaDesign. The two distributions over $\Delta ts_{\text{fold}}$ are shown in  Fig \ref{fig:seq_gen}(b-c). We also randomly pick some designed structures within the fold $i$ with $\Delta ts_{\text{fold}}(i) > 0.0$ and $\Delta ts_{\text{fold}}(i)< 0.0$, and visualize them in Fig.~\ref{fig:visual1} and Fig.~\ref{fig:visual2} in Section \ref{append: structure} in the appendix, respectively.

The computational efficiency in terms of inference is shown for Fold2Seq and RosettaDesign in Fig.~\ref{fig:seq_gen}(d) . Compared to RosettaDesign  on CPU, Fold2Seq on CPU and that on GPU are almost $100$ times and $5000$ times faster, respectively.

\vspace{-1em}
\paragraph{Generalizability Analysis.}
For each fold in the test sets, we calculate the maximum sequence identity (MSI) between a randomly selected sequence from that fold
and all folds in the training set (one random 
sequence per fold). We split all folds  in the test set into several bins. The performances of $ppl_{\text{fold}}$, $sr_{\text{fold}}$ and $cov_{\text{fold}}$ over bins are shown in Fig.~\ref{fig:dtm}.  In most cases, as the MSI increases, all methods have better performances in all three metrics except RosettaDesign which does not need a training set. For the low MSI bins  that demand generalizability, Fold2Seq is the best performer in $ppl_{\text{fold}}$ on the ID test set and in $cov_{\text{fold}}$ on both test sets, as well as the second best (next to Graph\_trans) in $ppl_{\text{fold}}$ on the OD test set and in $sr_{\text{fold}}$ on both test sets.
\vspace{-1em}
\paragraph{Ablation Study.}
To  rigorously delineate the contributions of each algorithmic innovation, we perform an ablation study 
(detailed in Section \ref{app:abl} in the appendix).
 The performance on the two test sets in terms of averaged  $sr_{\text{fold}}$ is summarized in Table \ref{tab:ablation}. \textbf{Key} observations are: (i)  `String' to `voxel' change and addition of 2 \textbf{FC} losses provide the largest performance gain (2-3\%). (ii) Use of transformer and  the cyclic loss improves performance by around 2\%.  (iii) In contrast, the improvement due to the addition of $\textbf{RE}_s$ and \textbf{CS} is minor. (iv) Further, the inclusion of  the two $\textbf{FC}$ losses as the intra-domain loss is crucial for joint embedding learning. 
 By calculating the averaged pairwise L2 distance among the hidden fold vectors, $h_f(\bm{y})$, for proteins in the OD test set, we found that such distance is smaller with FC losses (3.25) than without FC losses (5.35), which echoes our rationales of proposing fold-classification losses in the Method section. In summary, our novel design of the 3D voxel representation and the joint embedding learning framework, which includes  intra-domain and cyclic losses, leads to significant performance improvement.

\vspace{-0.5em}
\begin{table}[t!]
\caption{}
\vspace{-2mm}
\begin{subtable}[h]{0.45\textwidth}
\centering
\setlength{\arrayrulewidth}{0.5mm}
\caption{Avg. $sr_{\text{fold}}$ (std. dev.) (\%)  for variants in ablation study.}
\resizebox{0.75\textwidth}{!}{%
\begin{tabular}{r  c  c }
\hline
Model & ID Test & OD Test   \\
\hline
 cVAE & 18.2 (6.7) & 17.3 (5.2)\\
Trans\_string\_$\textbf{RE}_f$ &  20.0 (8.31) & 19.2 (3.45) \\
Trans\_voxel\_$\textbf{RE}_f$ & 22.5 (7.34)  &  21.3 (3.33)  \\    
+$\textbf{RE}_s$+$\textbf{CS}$ &  22.8 (8.01)   & 21.9 (2.34) \\
+2\textbf{FC}  &  25.6 (6.34)  &   23.7 (2.34) \\
+\textbf{CY} (Fold2Seq) &  \textbf{27.2} (\textbf{6.3}) & \textbf{25.2} (\textbf{3.2})  \\
\hline
\label{tab:ablation}
 \end{tabular}
}
\end{subtable}
\hspace{10em}
\begin{subtable}{0.45\textwidth}
\vspace{-0.7em}
\centering
\caption{Avg. $sr_{\text{structure}}$ (std. dev.) (\%)  for low resolution structures, and  Avg. $sr_{\text{fold}}$ and $cov_{\text{fold}}$ (std. dev.) (\%) for NMR ensembles.}
\setlength{\arrayrulewidth}{0.5mm}
\resizebox{0.73\textwidth}{!}{%
\begin{tabular}{r  c  c}
\hline
Model &  Low\_res Set &   \\
\hline
Graph\_trans  & 19.9 (4.8) &  \\
RosettaDesign &  17.2 (6.3) & \\
Fold2Seq &  \textbf{21.2 (3.1)} &   \\
\hline
$sr_{\text{fold}}$ for NMR &  ID Test &  OD Test  \\
\hline
$\text{Fold2Seq}_{\text{single}}$  &  24.1 (3.9)   &  22.2 (3.8) \\
$\text{Fold2Seq}_{\text{average}}$ &  25.2 (3.5)   &  24.1 (4.2) \\
\hline
$cov_{\text{fold}}$ for NMR &  ID Test  &  OD Test  \\
\hline
$\text{Graph\_trans}_{\text{all}}$  & 19.5 (26.3)  & 17.5 (28.5) \\
$\text{Fold2Seq}_{\text{single}}$  &  24.1 (3.9)   &  22.2 (3.8) \\
$\text{Fold2Seq}_{\text{average}}$ &  \textbf{25.2 (3.5)}   &  \textbf{24.1 (4.2)} \\ \hline 
 \end{tabular}
}
 \label{tab:lowres_nmr}
\end{subtable}
\vspace{\vslen}
\end{table}

\vspace{\vslen}

\section{Experiments on Real-world Challenges}
\vspace{\vslen}
To further explore the practical utility of our model, we perform three real-world  challenging design tasks conditioned on: (1) Low-resolution structures; (2) Structures with  missing residues; and  (3) NMR ensembles, representing low-quality, incomplete, and ambiguous data, respectively.  
\vspace{-1em}
\paragraph{Low-Resolution Structures.}
We first create the low-resolution structure dataset from Protein Data Bank, which contains 164 single-chain  proteins with low resolutions  ranging from 6\AA{} to 12\AA. This set has maximum sequence identity  (MSI) below 30\% compared to the training set.  We compare Fold2Seq's performance on this set with those of Graph\_trans  and RosettaDesign. Since the  fold information is not available for these low-resolution structures, we report structure-level sequence recovery ($sr_{\text{structure}}$) in Table \ref{tab:lowres_nmr},
showing that Fold2Seq outperforms other baselines.
As Fold2Seq uses the high-level fold information (by re-scaling the structure, discretizing the space, and smoothing the spatial secondary structure element information by neighborhood averaging), the model's performance  is  less sensitive compared to  RosettaDesign  or  Graph\_trans,  when  test structures are of lower resolution.
To further solidify the results, we  randomly pick three Fold2Seq's designed sequences on three different proteins respectively and recover their structures through  iTasser Suite. As a result, we receive $tr_\text{structure}: 0.39, 0.46, 0.33$ on protein 2W6G\_G, 5BW9\_L and 5UJ8\_H, respectively,  indicating structural similarity with the target structure.
\vspace{-1em}
\paragraph{Structures with Missing  Residues.}
We next perform the design task where the input structures have missing residues.  In order to mimic the real-world scenario, for every protein in our OD test set, we select a stretch of residues at random starting positions with length $p$, for which 1D and 3D information was removed. We compared Fold2Seq with Graph\_trans at $p=\{5\%, 10\%, 15\%, 20\%, 25\%, 30\%\}$, as shown in Fig.~\ref{fig:seq_gen}(e). When $p$ is small, the performance of  Fold2Seq is on par with Graph\_trans. As $p$ increases, Fold2Seq outperforms Graph\_trans with a consistent margin. We also perform one-sided, two-sample $t$-tests with null hypothesis: $sr_\text{fold}$ of Graph\_trans is larger than that of Fold2Seq and obtain P-value of 0.028 (at 10\% missing rate) or $<$1E-3 (at higher missing rates).  This shows that Fold2seq is less sensitive to the availability of complete and detailed backbone structure information.

\paragraph{NMR Structural Ensemble.}
We finally apply Fold2Seq to a structural ensemble  of NMR structures.  We filter the NMR structures from our two test sets and obtain 57 proteins in 30 folds from the ID set and 30  proteins in 10 folds from the OD set. On average each protein has around 20 structures. Handling NMR ensembles using Fold2Seq  is straightforward, when compared to Graph\_trans and RosettaDesign: after we obtain the voxel-based features through Eq \ref{eq:fold_features} for each model (structure) within one NMR ensemble,  we simply average them across all models. 
The sequence recovery results of Fold2Seq for NMR ensembles are shown in Table \ref{tab:lowres_nmr}, along with a single structure baseline. Results show that  Fold2Seq performs better on both ID and OD proteins, when ensemble structure information is available. This is consistent with our hypothesis that our fold representation better captures the structural variations present within a single fold. Moreover, we compare the coverage performance of Fold2Seq against multiple Graph\_trans which collectively uses the ensemble of all models within a NMR structure, and all NMR structures within a fold.  As shown in Table \ref{tab:lowres_nmr},  Fold2Seq designs using single or averaged SSE densities achieved higher coverage than Graph\_trans did using all structures, which shows that Fold2Seq has better efficiency and scalability for inverse fold design compared to structure-based methods with diverse structure inputs.  
 
\section{Conclusion and Future Work}
\vspace{\vslen}

In this paper, we design a novel transformer-based model to learn a fold representation  from scale-free and coarse topological features extracted from 3D voxels of secondary structure elements within and across folds and use  those as conditional inputs to design protein sequences. In order to mitigate the heterogeneity between the sequence domain and the fold domain, we learn the joint sequence--fold representation through novel intra-domain and cross-domain losses. On benchmark datasets containing single, high-resolution, complete input structures,   Fold2Seq performs better or similarly, compared to the existing neural net models and the state-of-the-art principle-driven RosettaDesign method, in terms of  perplexity, sequence recovery rate, coverage and structural recovery. 
Ablation study shows that this superior performance can be directly attributed to our novel algorithmic innovations, including the fold representation, joint sequence-fold embedding, and various losses. Moreover, we demonstrate the unique practical utility of Fold2Seq compared to structure-based neural net models  in a set of  real-world design tasks with challenging conditional inputs such as low resolution structures, structures with  region of missing residues,  and NMR structural ensembles. 

Future work will focus on upgrading fold embedding from convolutional neural networks to advanced architectures such as certain SE(3)-equivariant ones, learning representations in a continuous rather than a discrete fold space, and designing multi-domain and multi-chain proteins.

\section*{Acknowledgements} 
We thank IBM Research Internship Program for support.
The project was also in part supported by the National Institute of General Medical Sciences (R35GM124952 to YS).  Part of the computing resources was provided by the Texas A\&M High Performance Research Computing (HPRC).   

\bibliography{main}

\begin{thebibliography}{48}
\providecommand{\natexlab}[1]{#1}
\providecommand{\url}[1]{\texttt{#1}}
\expandafter\ifx\csname urlstyle\endcsname\relax
  \providecommand{\doi}[1]{doi: #1}\else
  \providecommand{\doi}{doi: \begingroup \urlstyle{rm}\Url}\fi

\bibitem[Arandjelovic \& Zisserman(2018)Arandjelovic and
  Zisserman]{arandjelovic2018objects}
Arandjelovic, R. and Zisserman, A.
\newblock Objects that sound.
\newblock In \emph{Proceedings of the European Conference on Computer Vision
  (ECCV)}, pp.\  435--451, 2018.

\bibitem[Basanta et~al.(2020)Basanta, Bick, Bera, Norn, Chow, Carter,
  Goreshnik, Dimaio, and Baker]{basanta2020enumerative}
Basanta, B., Bick, M.~J., Bera, A.~K., Norn, C., Chow, C.~M., Carter, L.~P.,
  Goreshnik, I., Dimaio, F., and Baker, D.
\newblock An enumerative algorithm for de novo design of proteins with diverse
  pocket structures.
\newblock \emph{Proceedings of the National Academy of Sciences}, 117\penalty0
  (36):\penalty0 22135--22145, 2020.

\bibitem[Berman et~al.(2000)Berman, Westbrook, Feng, Gilliland, Bhat, Weissig,
  Shindyalov, and Bourne]{berman2000protein}
Berman, H.~M., Westbrook, J., Feng, Z., Gilliland, G., Bhat, T.~N., Weissig,
  H., Shindyalov, I.~N., and Bourne, P.~E.
\newblock The protein data bank.
\newblock \emph{Nucleic acids research}, 28\penalty0 (1):\penalty0 235--242,
  2000.

\bibitem[Blundell et~al.(2015)Blundell, Cornebise, Kavukcuoglu, and
  Wierstra]{blundell2015weight}
Blundell, C., Cornebise, J., Kavukcuoglu, K., and Wierstra, D.
\newblock Weight uncertainty in neural network.
\newblock In \emph{International Conference on Machine Learning}, pp.\
  1613--1622. PMLR, 2015.

\bibitem[Boutemy et~al.(2011)Boutemy, King, Win, Hughes, Clarke, Blumenschein,
  Kamoun, and Banfield]{boutemy2011structures}
Boutemy, L.~S., King, S.~R., Win, J., Hughes, R.~K., Clarke, T.~A.,
  Blumenschein, T.~M., Kamoun, S., and Banfield, M.~J.
\newblock Structures of phytophthora rxlr effector proteins: a conserved but
  adaptable fold underpins functional diversity.
\newblock \emph{Journal of Biological Chemistry}, 286\penalty0 (41):\penalty0
  35834--35842, 2011.

\bibitem[Cao \& Shen(2021)Cao and Shen]{10.1093/bioinformatics/btab198}
Cao, Y. and Shen, Y.
\newblock {TALE: Transformer-based protein function Annotation with joint
  sequence–Label Embedding}.
\newblock \emph{Bioinformatics}, 03 2021.
\newblock ISSN 1367-4803.
\newblock \doi{10.1093/bioinformatics/btab198}.
\newblock btab198.

\bibitem[Chandra et~al.(2001)Chandra, Prabu, Suguna, and
  Vijayan]{chandra2001structural}
Chandra, N.~R., Prabu, M., Suguna, K., and Vijayan, M.
\newblock Structural similarity and functional diversity in proteins containing
  the legume lectin fold.
\newblock \emph{Protein Engineering}, 14\penalty0 (11):\penalty0 857--866,
  2001.

\bibitem[Chen et~al.(2018)Chen, Choy, Savva, Chang, Funkhouser, and
  Savarese]{chen2018text2shape}
Chen, K., Choy, C.~B., Savva, M., Chang, A.~X., Funkhouser, T., and Savarese,
  S.
\newblock Text2shape: Generating shapes from natural language by learning joint
  embeddings.
\newblock In \emph{Asian Conference on Computer Vision}, pp.\  100--116.
  Springer, 2018.

\bibitem[Chen et~al.(2019)Chen, Sun, Lin, Liu, Liu, Chong, Lu, Zhao, and
  Yang]{chen2019improve}
Chen, S., Sun, Z., Lin, L., Liu, Z., Liu, X., Chong, Y., Lu, Y., Zhao, H., and
  Yang, Y.
\newblock To improve protein sequence profile prediction through image
  captioning on pairwise residue distance map.
\newblock \emph{Journal of chemical information and modeling}, 60\penalty0
  (1):\penalty0 391--399, 2019.

\bibitem[Das et~al.(2018)Das, Wadhawan, Chang, Sercu, Santos, Riemer,
  Chenthamarakshan, Padhi, and Mojsilovic]{das2018pepcvae}
Das, P., Wadhawan, K., Chang, O., Sercu, T., Santos, C.~D., Riemer, M.,
  Chenthamarakshan, V., Padhi, I., and Mojsilovic, A.
\newblock Pepcvae: Semi-supervised targeted design of antimicrobial peptide
  sequences.
\newblock \emph{arXiv preprint arXiv:1810.07743}, 2018.

\bibitem[David et~al.(2020)David, Thakkar, Mercado, and
  Engkvist]{david2020molecular}
David, L., Thakkar, A., Mercado, R., and Engkvist, O.
\newblock Molecular representations in ai-driven drug discovery: a review and
  practical guide.
\newblock \emph{Journal of Cheminformatics}, 12\penalty0 (1):\penalty0 1--22,
  2020.

\bibitem[Dognin et~al.(2019)Dognin, Melnyk, Mroueh, Ross, and
  Sercu]{dognin2019adversarial}
Dognin, P., Melnyk, I., Mroueh, Y., Ross, J., and Sercu, T.
\newblock Adversarial semantic alignment for improved image captions.
\newblock In \emph{Proceedings of the IEEE Conference on Computer Vision and
  Pattern Recognition}, pp.\  10463--10471, 2019.

\bibitem[Fan et~al.(2018)Fan, Lewis, and Dauphin]{fan2018hierarchical}
Fan, A., Lewis, M., and Dauphin, Y.
\newblock Hierarchical neural story generation.
\newblock \emph{arXiv preprint arXiv:1805.04833}, 2018.

\bibitem[Greener et~al.(2018)Greener, Moffat, and Jones]{greener2018design}
Greener, J.~G., Moffat, L., and Jones, D.~T.
\newblock Design of metalloproteins and novel protein folds using variational
  autoencoders.
\newblock \emph{Scientific reports}, 8\penalty0 (1):\penalty0 1--12, 2018.

\bibitem[Gu et~al.(2018)Gu, Cai, Joty, Niu, and Wang]{gu2018look}
Gu, J., Cai, J., Joty, S.~R., Niu, L., and Wang, G.
\newblock Look, imagine and match: Improving textual-visual cross-modal
  retrieval with generative models.
\newblock In \emph{Proceedings of the IEEE Conference on Computer Vision and
  Pattern Recognition}, pp.\  7181--7189, 2018.

\bibitem[Hou et~al.(2003)Hou, Sims, Zhang, and Kim]{hou2003global}
Hou, J., Sims, G.~E., Zhang, C., and Kim, S.-H.
\newblock A global representation of the protein fold space.
\newblock \emph{Proceedings of the National Academy of Sciences}, 100\penalty0
  (5):\penalty0 2386--2390, 2003.

\bibitem[Huang et~al.(2011)Huang, Ban, Richter, Andre, Vernon, Schief, and
  Baker]{huang2011rosettaremodel}
Huang, P.-S., Ban, Y.-E.~A., Richter, F., Andre, I., Vernon, R., Schief, W.~R.,
  and Baker, D.
\newblock Rosettaremodel: a generalized framework for flexible backbone protein
  design.
\newblock \emph{PloS one}, 6\penalty0 (8):\penalty0 e24109, 2011.

\bibitem[Huang et~al.(2016)Huang, Boyken, and Baker]{huang2016coming}
Huang, P.-S., Boyken, S.~E., and Baker, D.
\newblock The coming of age of de novo protein design.
\newblock \emph{Nature}, 537\penalty0 (7620):\penalty0 320--327, 2016.

\bibitem[Ingraham et~al.(2019)Ingraham, Garg, Barzilay, and
  Jaakkola]{ingraham2019generative}
Ingraham, J., Garg, V., Barzilay, R., and Jaakkola, T.
\newblock Generative models for graph-based protein design.
\newblock In \emph{Advances in Neural Information Processing Systems}, pp.\
  15820--15831, 2019.

\bibitem[Kabsch \& Sander(1983)Kabsch and Sander]{kabsch1983dictionary}
Kabsch, W. and Sander, C.
\newblock Dictionary of protein secondary structure: pattern recognition of
  hydrogen-bonded and geometrical features.
\newblock \emph{Biopolymers: Original Research on Biomolecules}, 22\penalty0
  (12):\penalty0 2577--2637, 1983.

\bibitem[Karimi et~al.(2020)Karimi, Zhu, Cao, and Shen]{karimi2020novo}
Karimi, M., Zhu, S., Cao, Y., and Shen, Y.
\newblock De novo protein design for novel folds using guided conditional
  wasserstein generative adversarial networks.
\newblock \emph{Journal of Chemical Information and Modeling}, 60\penalty0
  (12):\penalty0 5667--5681, 2020.

\bibitem[Koga et~al.(2012)Koga, Tatsumi-Koga, Liu, Xiao, Acton, Montelione, and
  Baker]{koga2012principles}
Koga, N., Tatsumi-Koga, R., Liu, G., Xiao, R., Acton, T.~B., Montelione, G.~T.,
  and Baker, D.
\newblock Principles for designing ideal protein structures.
\newblock \emph{Nature}, 491\penalty0 (7423):\penalty0 222--227, 2012.

\bibitem[Kraemer-Pecore et~al.(2001)Kraemer-Pecore, Wollacott, and
  Desjarlais]{kraemer2001computational}
Kraemer-Pecore, C.~M., Wollacott, A.~M., and Desjarlais, J.~R.
\newblock Computational protein design.
\newblock \emph{Current Opinion in Chemical Biology}, 5\penalty0 (6):\penalty0
  690--695, 2001.

\bibitem[Kuhlman et~al.(2003)Kuhlman, Dantas, Ireton, Varani, Stoddard, and
  Baker]{kuhlman2003design}
Kuhlman, B., Dantas, G., Ireton, G.~C., Varani, G., Stoddard, B.~L., and Baker,
  D.
\newblock Design of a novel globular protein fold with atomic-level accuracy.
\newblock \emph{science}, 302\penalty0 (5649):\penalty0 1364--1368, 2003.

\bibitem[Madani et~al.(2020)Madani, McCann, Naik, Keskar, Anand, Eguchi, Huang,
  and Socher]{madani2020progen}
Madani, A., McCann, B., Naik, N., Keskar, N.~S., Anand, N., Eguchi, R.~R.,
  Huang, P.-S., and Socher, R.
\newblock Progen: Language modeling for protein generation.
\newblock \emph{arXiv preprint arXiv:2004.03497}, 2020.

\bibitem[Murzin et~al.(1995)Murzin, Brenner, Hubbard, Chothia,
  et~al.]{murzin1995scop}
Murzin, A.~G., Brenner, S.~E., Hubbard, T., Chothia, C., et~al.
\newblock Scop: a structural classification of proteins database for the
  investigation of sequences and structures.
\newblock \emph{Journal of molecular biology}, 247\penalty0 (4):\penalty0
  536--540, 1995.

\bibitem[Needleman \& Wunsch(1970)Needleman and Wunsch]{needleman1970general}
Needleman, S.~B. and Wunsch, C.~D.
\newblock A general method applicable to the search for similarities in the
  amino acid sequence of two proteins.
\newblock \emph{Journal of molecular biology}, 48\penalty0 (3):\penalty0
  443--453, 1970.

\bibitem[Ngiam et~al.(2011)Ngiam, Khosla, Kim, Nam, Lee, and
  Ng]{ngiam2011multimodal}
Ngiam, J., Khosla, A., Kim, M., Nam, J., Lee, H., and Ng, A.~Y.
\newblock Multimodal deep learning.
\newblock In \emph{ICML}, 2011.

\bibitem[O'Connell et~al.(2018)O'Connell, Li, Hanson, Heffernan, Lyons,
  Paliwal, Dehzangi, Yang, and Zhou]{o2018spin2}
O'Connell, J., Li, Z., Hanson, J., Heffernan, R., Lyons, J., Paliwal, K.,
  Dehzangi, A., Yang, Y., and Zhou, Y.
\newblock Spin2: Predicting sequence profiles from protein structures using
  deep neural networks.
\newblock \emph{Proteins: Structure, Function, and Bioinformatics}, 86\penalty0
  (6):\penalty0 629--633, 2018.

\bibitem[Orengo et~al.(1997)Orengo, Michie, Jones, Jones, Swindells, and
  Thornton]{orengo1997cath}
Orengo, C.~A., Michie, A.~D., Jones, S., Jones, D.~T., Swindells, M.~B., and
  Thornton, J.~M.
\newblock Cath--a hierarchic classification of protein domain structures.
\newblock \emph{Structure}, 5\penalty0 (8):\penalty0 1093--1109, 1997.

\bibitem[Paszke et~al.(2019)Paszke, Gross, Massa, Lerer, Bradbury, Chanan,
  Killeen, Lin, Gimelshein, Antiga, et~al.]{paszke2019pytorch}
Paszke, A., Gross, S., Massa, F., Lerer, A., Bradbury, J., Chanan, G., Killeen,
  T., Lin, Z., Gimelshein, N., Antiga, L., et~al.
\newblock Pytorch: An imperative style, high-performance deep learning library.
\newblock In \emph{Advances in neural information processing systems}, pp.\
  8026--8037, 2019.

\bibitem[Peng \& Qi(2019)Peng and Qi]{peng2019cm}
Peng, Y. and Qi, J.
\newblock {CM-GANs}: Cross-modal generative adversarial networks for common
  representation learning.
\newblock \emph{ACM Transactions on Multimedia Computing, Communications, and
  Applications (TOMM)}, 15\penalty0 (1):\penalty0 1--24, 2019.

\bibitem[Rost(1999)]{rost1999twilight}
Rost, B.
\newblock Twilight zone of protein sequence alignments.
\newblock \emph{Protein engineering}, 12\penalty0 (2):\penalty0 85--94, 1999.

\bibitem[Sillitoe et~al.(2019)Sillitoe, Dawson, Lewis, Das, Lees, Ashford,
  Tolulope, Scholes, Senatorov, Bujan, et~al.]{sillitoe2019cath}
Sillitoe, I., Dawson, N., Lewis, T.~E., Das, S., Lees, J.~G., Ashford, P.,
  Tolulope, A., Scholes, H.~M., Senatorov, I., Bujan, A., et~al.
\newblock Cath: expanding the horizons of structure-based functional
  annotations for genome sequences.
\newblock \emph{Nucleic acids research}, 47\penalty0 (D1):\penalty0 D280--D284,
  2019.

\bibitem[Socher et~al.(2013)Socher, Ganjoo, Manning, and Ng]{socher2013zero}
Socher, R., Ganjoo, M., Manning, C.~D., and Ng, A.
\newblock Zero-shot learning through cross-modal transfer.
\newblock In \emph{Advances in neural information processing systems}, pp.\
  935--943, 2013.

\bibitem[Strokach et~al.(2020)Strokach, Becerra, Corbi-Verge, Perez-Riba, and
  Kim]{strokach2020fast}
Strokach, A., Becerra, D., Corbi-Verge, C., Perez-Riba, A., and Kim, P.~M.
\newblock Fast and flexible protein design using deep graph neural networks.
\newblock \emph{Cell Systems}, 11\penalty0 (4):\penalty0 402--411.e4, 2020.
\newblock ISSN 2405-4712.
\newblock \doi{https://doi.org/10.1016/j.cels.2020.08.016}.
\newblock URL
  \url{https://www.sciencedirect.com/science/article/pii/S2405471220303276}.

\bibitem[Suzek et~al.(2015)Suzek, Wang, Huang, McGarvey, Wu, and
  Consortium]{suzek2015uniref}
Suzek, B.~E., Wang, Y., Huang, H., McGarvey, P.~B., Wu, C.~H., and Consortium,
  U.
\newblock Uniref clusters: a comprehensive and scalable alternative for
  improving sequence similarity searches.
\newblock \emph{Bioinformatics}, 31\penalty0 (6):\penalty0 926--932, 2015.

\bibitem[Taylor(2002)]{taylor2002periodic}
Taylor, W.~R.
\newblock A ‘periodic table’for protein structures.
\newblock \emph{Nature}, 416\penalty0 (6881):\penalty0 657--660, 2002.

\bibitem[Toutanova et~al.(2015)Toutanova, Chen, Pantel, Poon, Choudhury, and
  Gamon]{toutanova2015representing}
Toutanova, K., Chen, D., Pantel, P., Poon, H., Choudhury, P., and Gamon, M.
\newblock Representing text for joint embedding of text and knowledge bases.
\newblock In \emph{Proceedings of the 2015 conference on empirical methods in
  natural language processing}, pp.\  1499--1509, 2015.

\bibitem[Vaswani et~al.(2017)Vaswani, Shazeer, Parmar, Uszkoreit, Jones, Gomez,
  Kaiser, and Polosukhin]{vaswani2017attention}
Vaswani, A., Shazeer, N., Parmar, N., Uszkoreit, J., Jones, L., Gomez, A.~N.,
  Kaiser, {\L}., and Polosukhin, I.
\newblock Attention is all you need.
\newblock In \emph{Advances in neural information processing systems}, pp.\
  5998--6008, 2017.

\bibitem[Wang et~al.(2018{\natexlab{a}})Wang, Li, Wang, Zhang, Shen, Zhang,
  Henao, and Carin]{wang2018joint}
Wang, G., Li, C., Wang, W., Zhang, Y., Shen, D., Zhang, X., Henao, R., and
  Carin, L.
\newblock Joint embedding of words and labels for text classification.
\newblock \emph{arXiv preprint arXiv:1805.04174}, 2018{\natexlab{a}}.

\bibitem[Wang et~al.(2018{\natexlab{b}})Wang, Cao, Zhang, and
  Qi]{wang2018computational}
Wang, J., Cao, H., Zhang, J.~Z., and Qi, Y.
\newblock Computational protein design with deep learning neural networks.
\newblock \emph{Scientific reports}, 8\penalty0 (1):\penalty0 1--9,
  2018{\natexlab{b}}.

\bibitem[Wang et~al.(2013)Wang, He, Wang, Wang, and Tan]{wang2013learning}
Wang, K., He, R., Wang, W., Wang, L., and Tan, T.
\newblock Learning coupled feature spaces for cross-modal matching.
\newblock In \emph{Proceedings of the IEEE International Conference on Computer
  Vision}, pp.\  2088--2095, 2013.

\bibitem[Woolfson et~al.(2015)Woolfson, Bartlett, Burton, Heal, Niitsu,
  Thomson, and Wood]{woolfson2015novo}
Woolfson, D.~N., Bartlett, G.~J., Burton, A.~J., Heal, J.~W., Niitsu, A.,
  Thomson, A.~R., and Wood, C.~W.
\newblock De novo protein design: how do we expand into the universe of
  possible protein structures?
\newblock \emph{Current opinion in structural biology}, 33:\penalty0 16--26,
  2015.

\bibitem[Yang et~al.(2015)Yang, Yan, Roy, Xu, Poisson, and
  Zhang]{yang2015tasser}
Yang, J., Yan, R., Roy, A., Xu, D., Poisson, J., and Zhang, Y.
\newblock The i-tasser suite: protein structure and function prediction.
\newblock \emph{Nature methods}, 12\penalty0 (1):\penalty0 7--8, 2015.

\bibitem[Zhang \& Skolnick(2004)Zhang and Skolnick]{zhang2004scoring}
Zhang, Y. and Skolnick, J.
\newblock Scoring function for automated assessment of protein structure
  template quality.
\newblock \emph{Proteins: Structure, Function, and Bioinformatics}, 57\penalty0
  (4):\penalty0 702--710, 2004.

\bibitem[Zhang \& Saligrama(2016)Zhang and Saligrama]{zhang2016zero}
Zhang, Z. and Saligrama, V.
\newblock Zero-shot learning via joint latent similarity embedding.
\newblock In \emph{proceedings of the IEEE Conference on Computer Vision and
  Pattern Recognition}, pp.\  6034--6042, 2016.

\bibitem[Zhu et~al.(2017)Zhu, Park, Isola, and Efros]{zhu2017unpaired}
Zhu, J.-Y., Park, T., Isola, P., and Efros, A.~A.
\newblock Unpaired image-to-image translation using cycle-consistent
  adversarial networks.
\newblock In \emph{Proceedings of the IEEE international conference on computer
  vision}, pp.\  2223--2232, 2017.

\end{thebibliography}
\bibliographystyle{icml2021}

\clearpage
\begin{appendices}
\section{3D Extension of the Sinusoidal Positional Encoding}
\label{append:3d positional encoding}
We use a simple extension of the sinusoidal encoding described in the original transformer model~\citep{vaswani2017attention} to encode each position in our Structure Encoder. 
\begin{equation}
\begin{aligned}
    \text{PE}(x,y,z,2i) &= \sin(x/10000^{2i/h})\\
                        &+\sin(y/10000^{2i/h})\\
                        &+\sin(z/10000^{2i/h}) \\
     \text{PE}(x,y,z,2i+1) &= \cos(x/10000^{2i/h})\\
                           &+\cos(y/10000^{2i/h})\\
                           &+ \cos(z/10000^{2i/h})    
\end{aligned}
\end{equation}

\section{Comparison between Two Training Strategies} \label{append: loss}

In this section, we compare the performance between one-stage training  and two-stage training strategies.  In the one-stage strategy, we train our model through the 5 loss terms in Eq (4) together. While in the two-stage strategy, we first train our model using $L_1$  and then train using $L_2$. 

We first compare the learning curves. As the $\textbf{RE}_f$ loss represents the quality of the model, we plot the $\textbf{RE}_f$ loss vs epochs on both training and validation sets.

\begin{figure}[!hbt]
    \centering
    \includegraphics[width=0.47\textwidth]{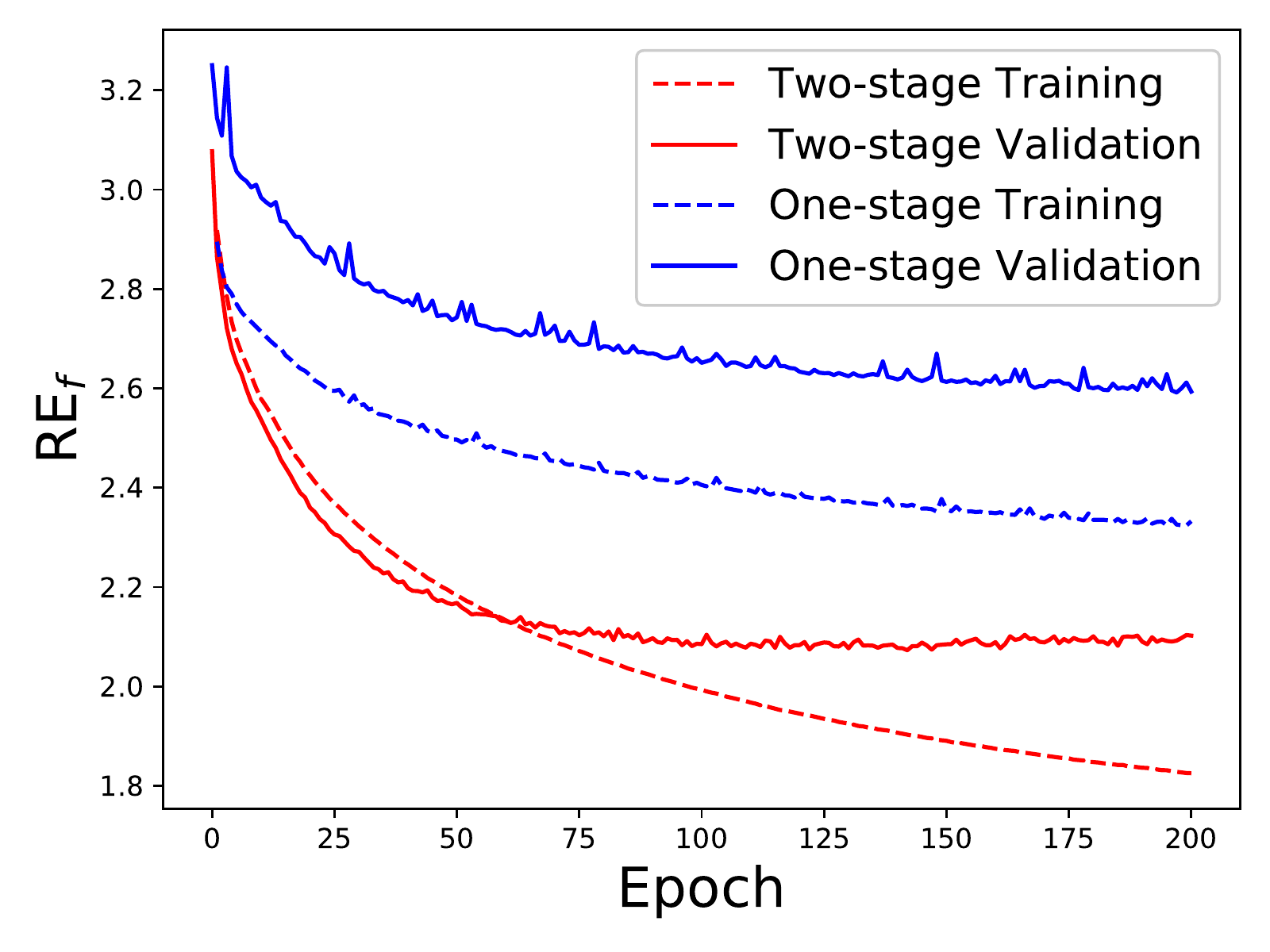}
    \caption{The fold2seq loss($\textbf{RE}_f$) curves of two training strategies on training and validation set.  }
    \label{fig:loss}
\end{figure}

\begin{table*}[t!]
\caption{Performance of two training strategies assessed in the sequence domain. }
\begin{subtable}[h]{0.45\textwidth}
\setlength{\arrayrulewidth}{0.5mm}
\caption{Avg. $ppl_{\text{fold}}(i)$ (std. dev.) (\%).}
\begin{tabular}{r  c  c }
\hline
Model & ID Test & OD Test \\
\hline
Uniform & 20.0 & 20.0  \\
Natural & 18.0 & 18.0 \\ 
\hline
One-stage strategy &  13.1 (4.3) & 15.3 (3.2) \\
Two-stage strategy & \textbf{9.0} (5.3) & 12.0 (2.4)  \\ \hline
 \end{tabular}
\label{tab:perplexity}
\end{subtable}
\begin{subtable}[h]{0.45\textwidth}
\centering
\caption{Avg. $sr_{\text{fold}}(i)$ (std. dev.) (\%).}
\setlength{\arrayrulewidth}{0.5mm}
\begin{tabular}{r   c  c }
\hline
Model & ID Test & OD Test   \\
\hline
Random across two  folds  & 12.8 (7.9) & 12.8 (7.9) \\
\hline
One-stage strategy & 22.2 (4.3) & 20.3 (3.2) \\
Two-stage strategy &  \textbf{27.2 (6.3)}  &   \textbf{25.2 (3.2)}\\
\hline
Random within same fold & 39.1 (9.4)  & 39.1 (9.4) \\         \hline   
 \end{tabular}
\end{subtable}
\label{tab:two training strategies}
\end{table*}

As shown in Fig \ref{fig:loss}, the two-stage strategy significantly outperforms the one-stage strategy.  To further demonstrate this point, we calculate the per-residue perplexity and the average sequence recovery rate on the two test sets. As shown in Table \ref{tab:two training strategies}, the same conclusion can be drawn. These results  validate our design choice of a two-stage training strategy.

\section{Dataset Statistics}\label{append:datasets}
The statistics of our various datasets are given below.
\begin{itemize}
    \item Training set includes 45995 proteins belonging to a total of 971 folds.
    \item Validation set includes 4159 proteins belonging to a total of 185 folds.
    \item In-distribution (ID) test set includes 1131 proteins belonging to  a total of 181 folds.
    \item Out-of-distribution (OD) test set includes 203 proteins belonging to a total of 27  folds. 
\end{itemize}

\section{Sequence Identity Measurement} \label{app:seq_id}
Sequence identity is measured through the Needleman--Wunsch algorithm \citep{needleman1970general} with the Blossum62 scoring matrix.
\section{Structure Level Performance Metrics} \label{app:str_level}

While we reported fold level performance metrics in the main paper, we also report the corresponding structure level metrics below. Fold2Seq outperforms all other methods except Graph\_trans in $ppl_\text{structure}(i)$ and $sr_{\text{structure}}(i)$. Note that Graph\_trans has an inherent advantage here because it uses the entire structure, where as Fold2Seq only uses high level fold information. Similar to fold based metrics, Fold2Seq performs better in the Missing Residues experiment and can also handle NMR Structural Ensembles.
\begin{table*}[!hbt]
\caption{Performance of different methods assessed in the sequence domain. }
\vspace{-2mm}
\begin{subtable}[h]{0.45\textwidth}
\setlength{\arrayrulewidth}{0.5mm}
\caption{Avg. $ppl_\text{structure}(i)$ (std. dev.)}
\begin{tabular}{r  c  c }
\hline
Model & ID Test & OD Test \\
\hline
Uniform & 20.0 & 20.0  \\
Natural & 18.0 & 18.0 \\ 
\hline
 cVAE & 14.8 & 16.3  \\
gcWGAN &  13.5 & 15.2 \\
Graph\_trans & \textbf{7.3}   &  \textbf{10.3} \\
Fold2Seq & 8.1 &  11.9  \\ \hline 
 \end{tabular}
\label{tab:perplexity}
\end{subtable}
\hspace{-2em}
\begin{subtable}[h]{0.45\textwidth}
\centering
\caption{Avg. $sr_{\text{structure}}(i)$ (std. dev.) (\%).}
\setlength{\arrayrulewidth}{0.5mm}
\begin{tabular}{r   c  c }
\hline
Model & ID Test & OD Test   \\
\hline
Random across two  folds  & 12.8 (7.94) & 12.8 (7.94) \\
\hline
cVAE & 17.7 (7.34) & 15.3 (5.34) \\
gcWGAN &  17.5 (6.35) & 14.1 (3.45) \\
RosettaDesign & 20.3 (5.13) &  20.2 (2.98) \\   Graph\_trans & \textbf{29.3 (4.3)} & \textbf{27.4 (3.2)} \\
Fold2Seq & 27.1 (6.31) & {24.1} ({2.64})  \\
\hline
Random within same fold & 39.1 (9.35)  & 39.1 (9.35) \\          \hline
 \end{tabular}
\label{tab:seq_recover_mean}
\end{subtable}
\vspace{-2mm}
\end{table*}

\begin{figure*}[!hbt]
    \centering
    \begin{subfigure}[b]{0.45\textwidth}
     \includegraphics[width=\textwidth]{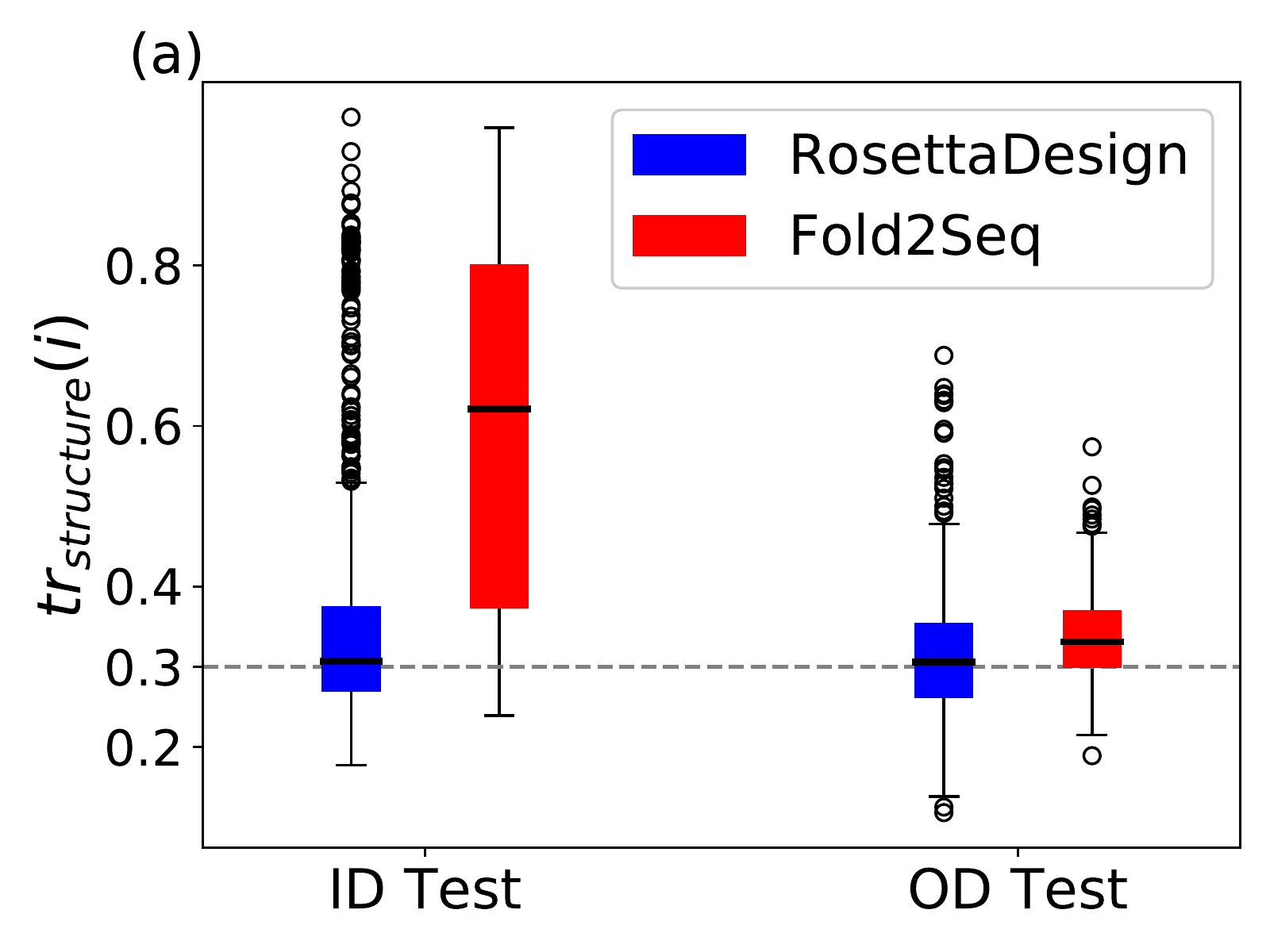}
     \end{subfigure}
        \begin{subfigure}[b]{0.45\textwidth}
    \includegraphics[width=\textwidth]{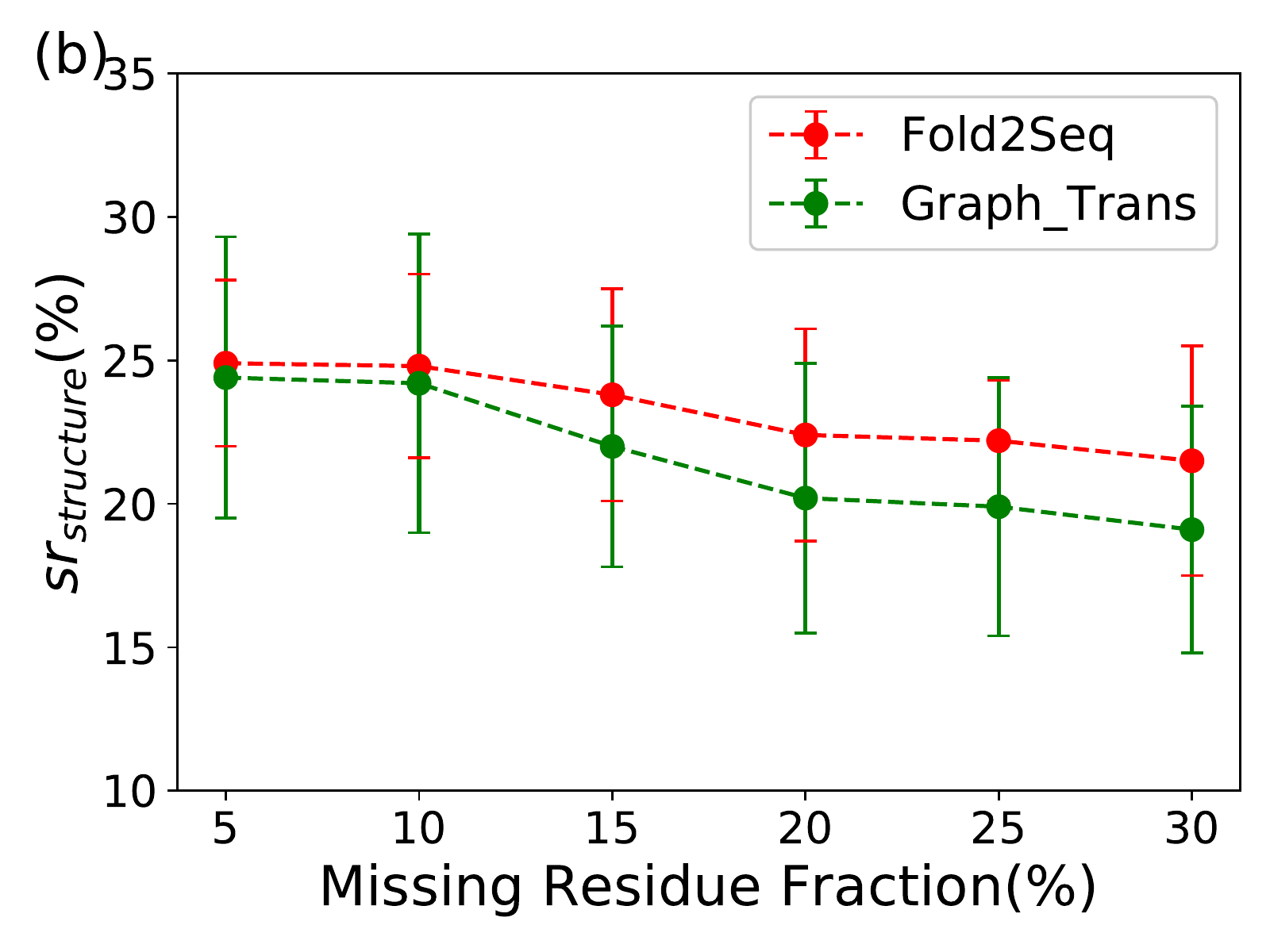}
    \end{subfigure}
    \vspace{-4mm}
        \caption{(a). $tr_{\text{structure}}(i)$ distributions  of RosettaDesign and Fold2Seq.  (c). Avg. $sr_{\text{structure}}$ (\%) for the OD test set  with a string of missing residues.}
            \label{fig:seq_gen}
          
\end{figure*}

\begin{table}[H]
\caption{Avg. $sr_{\text{structure}}(i)$ (std. dev.) (\%). for  NMR ensemble. }
\setlength{\arrayrulewidth}{0.5mm}
\begin{center}
\begin{tabular}{r  c  c }
\hline
NMR Input &  ID  &  OD   \\
\hline
Single  &  22.7 (3.4)   &  20.9 (4.2) \\
Ensemble &  24.1 (4.6)   &  22.3 (3.1) \\
\hline
 \end{tabular}
 \end{center}
\end{table}

\section{Generalizability Analysis of Performances} \label{append:dtm}

\begin{figure*}[!h]
    \centering
    \begin{subfigure}[b]{0.32\textwidth}
    \includegraphics[width=\textwidth]{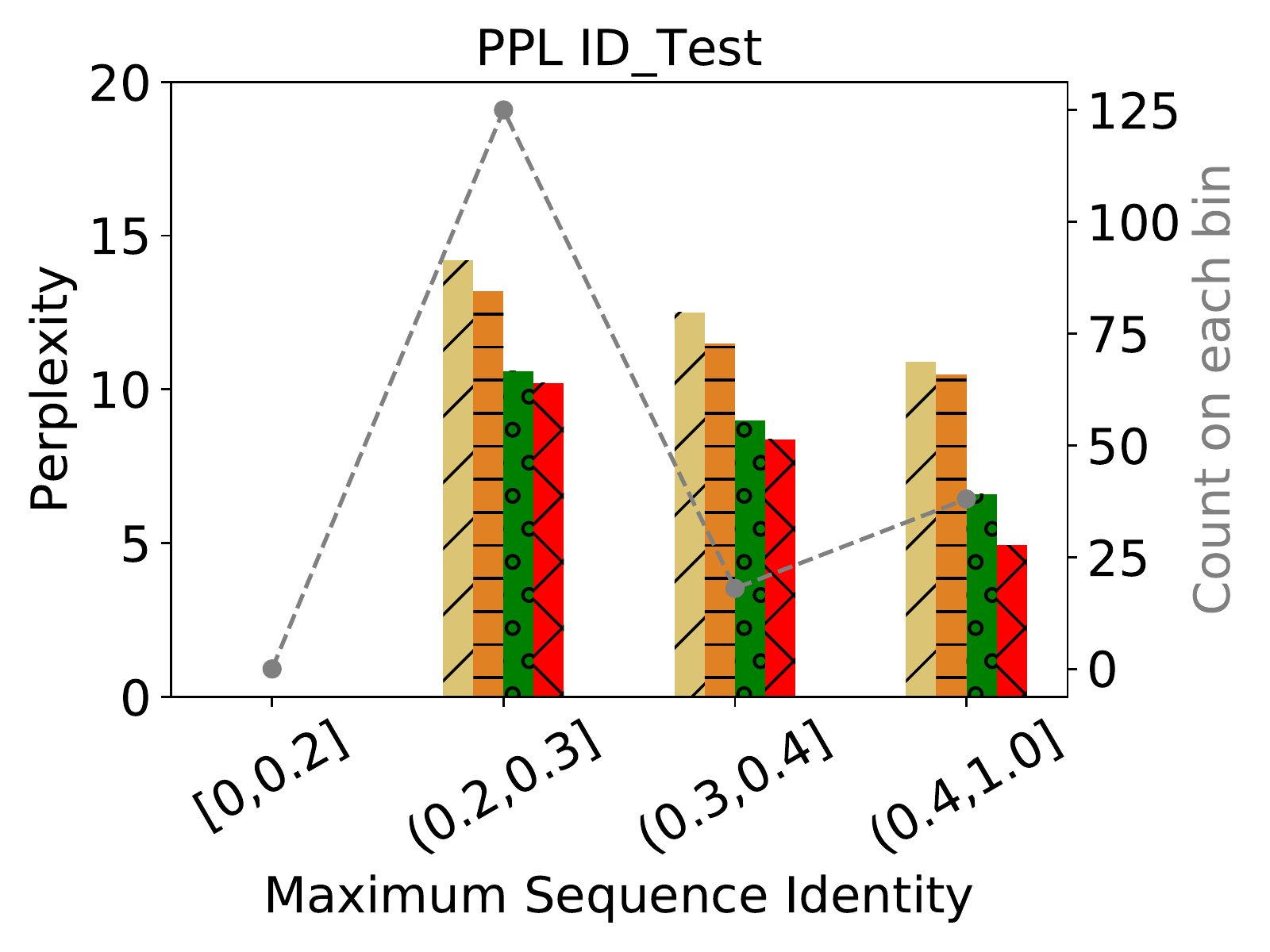}
    \end{subfigure}
        \begin{subfigure}[b]{0.32\textwidth}
    \includegraphics[width=\textwidth]{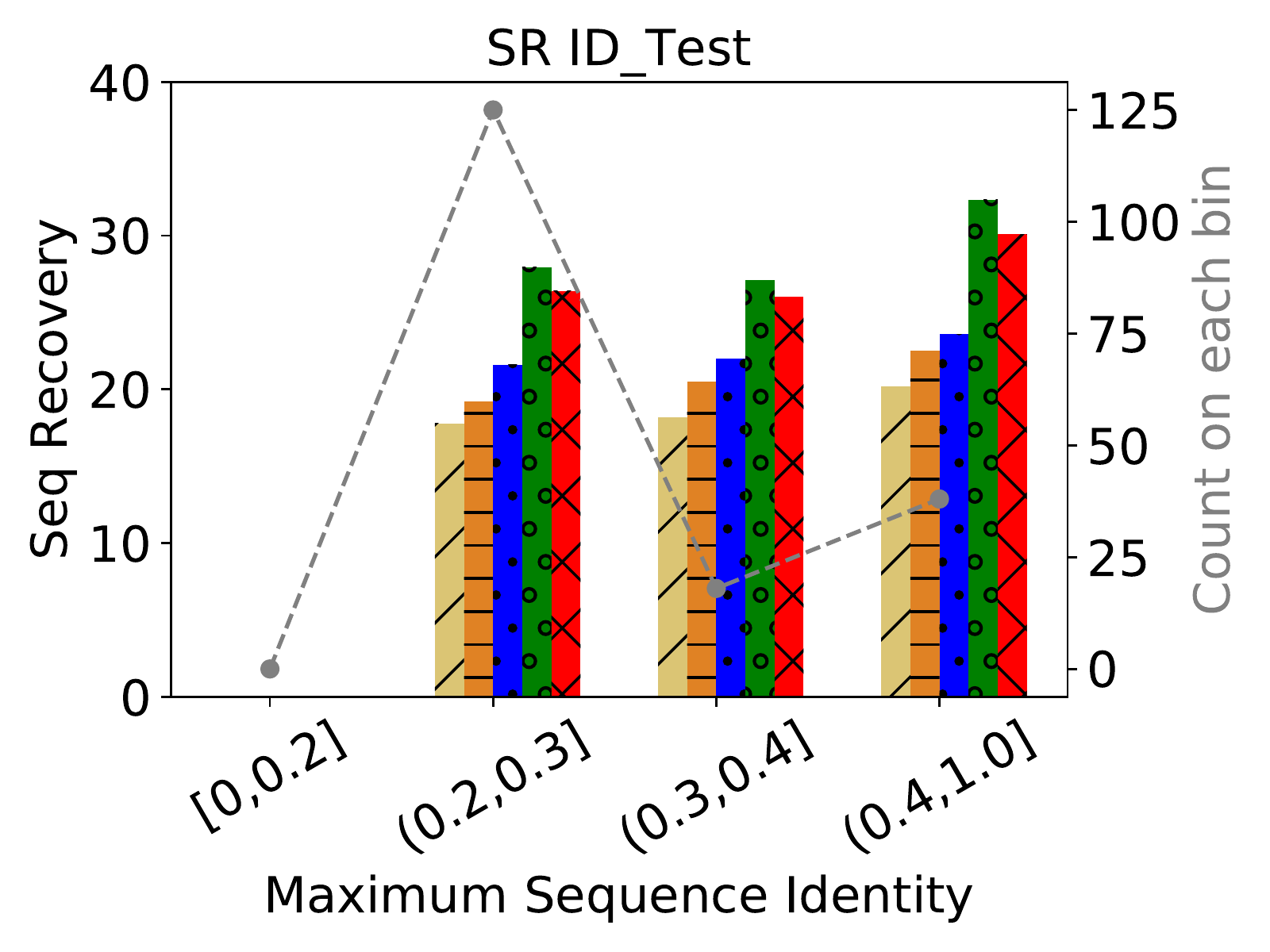}
     \end{subfigure}
         \begin{subfigure}[b]{0.32\textwidth}
    \includegraphics[width=\textwidth]{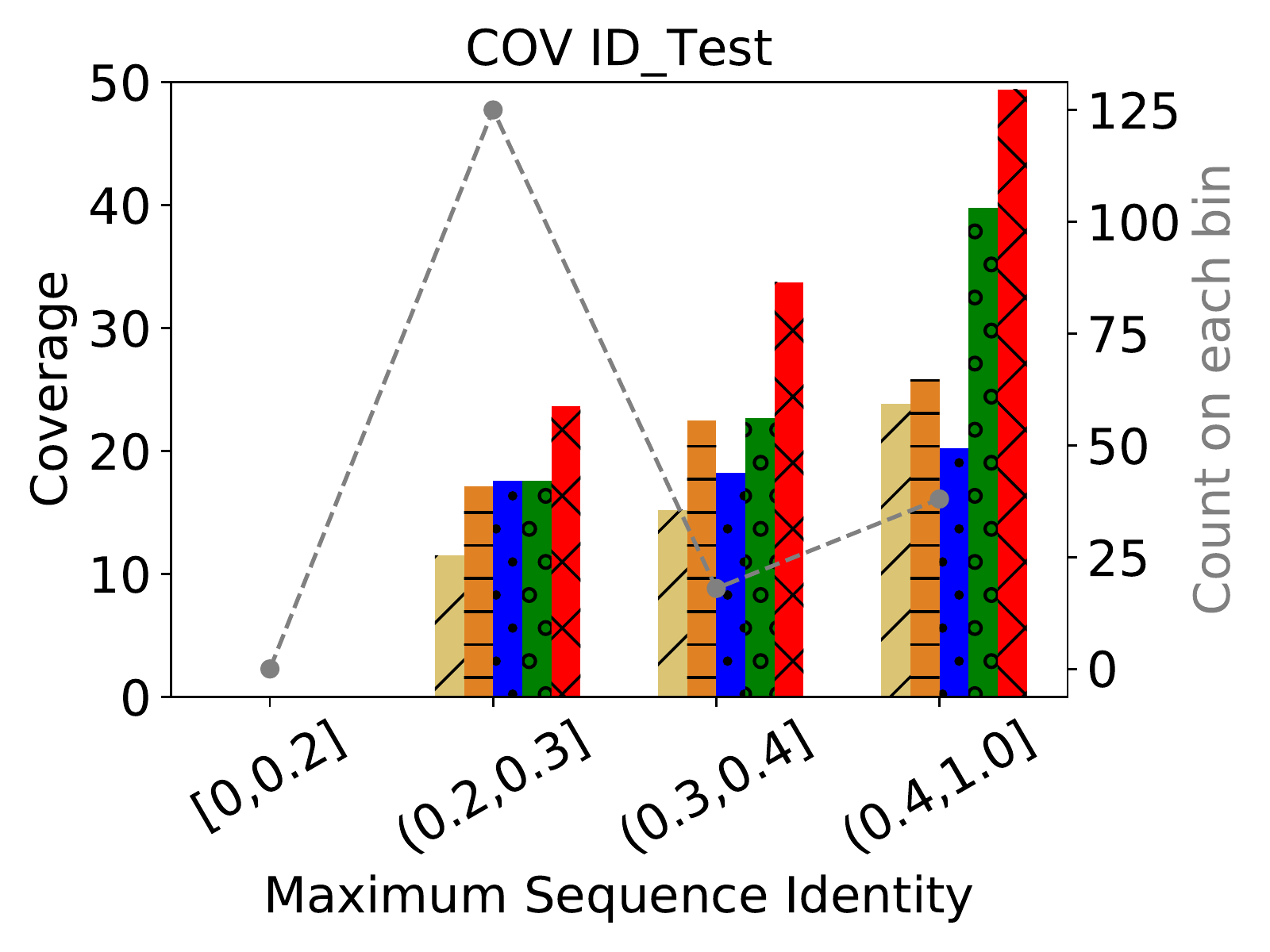}
    \end{subfigure}
        \begin{subfigure}[b]{0.32\textwidth}
    \includegraphics[width=\textwidth]{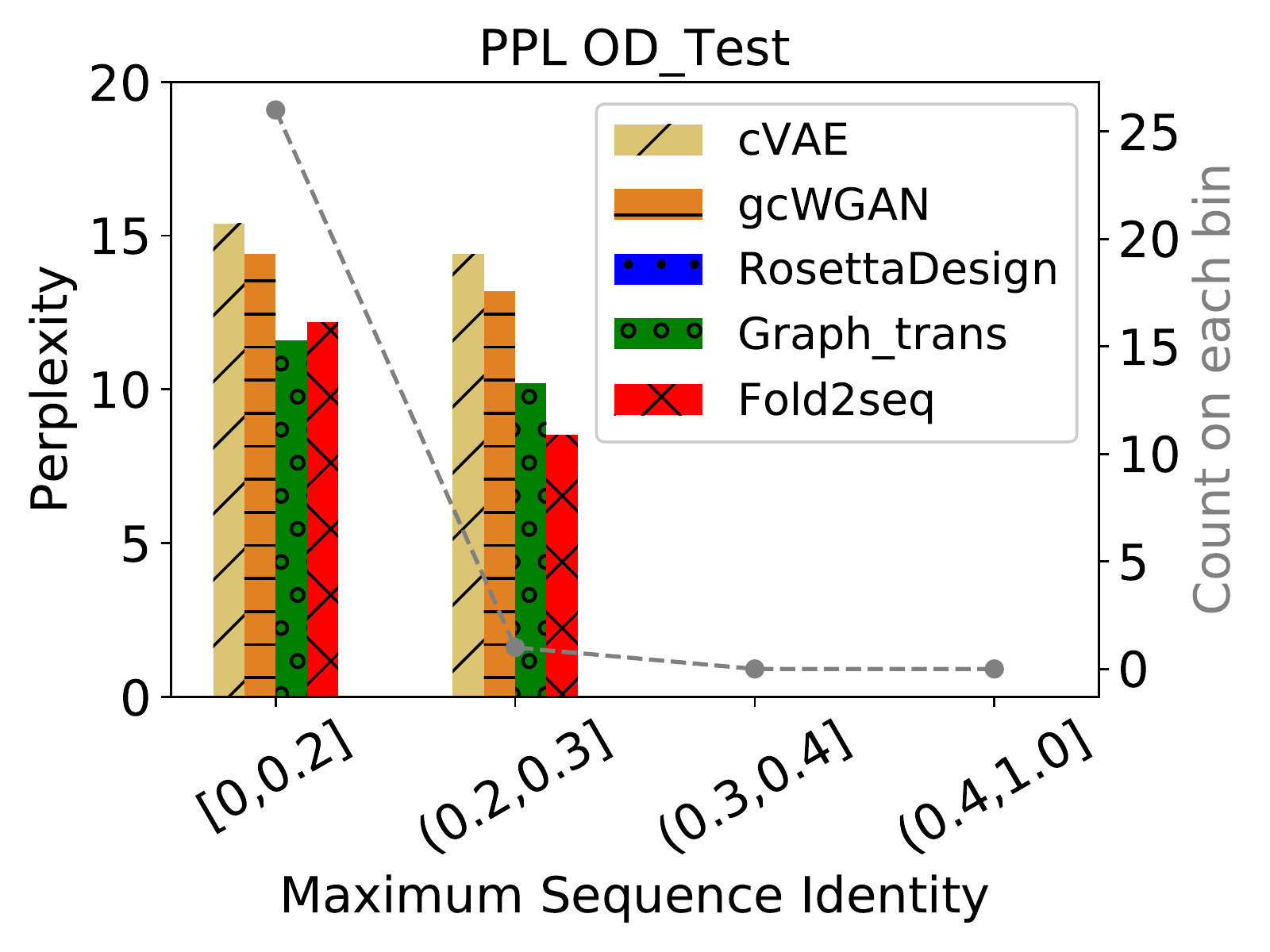}
     \end{subfigure}
         \begin{subfigure}[b]{0.32\textwidth}
    \includegraphics[width=\textwidth]{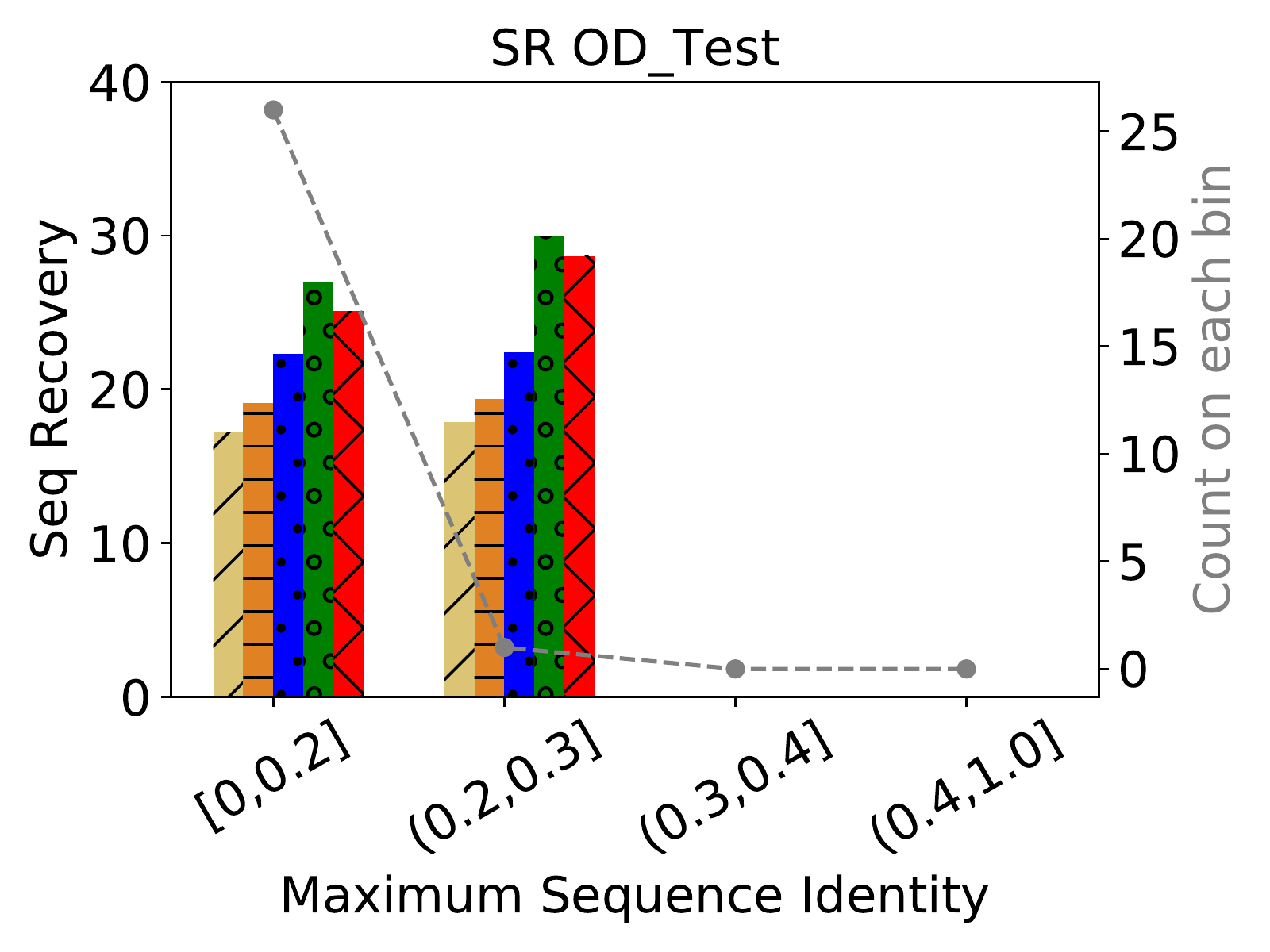}
    \end{subfigure}
        \begin{subfigure}[b]{0.32\textwidth}
    \includegraphics[width=\textwidth]{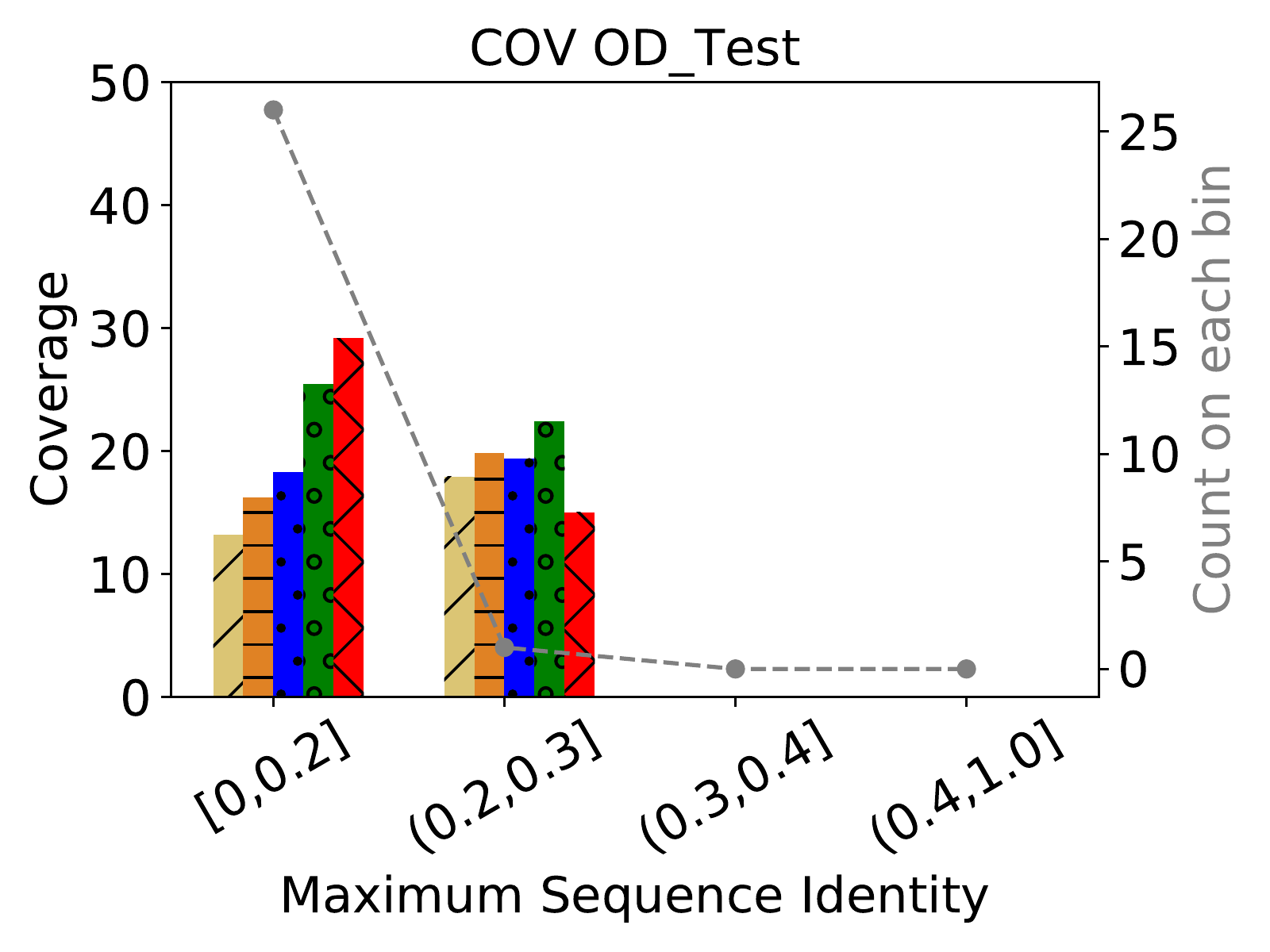}
     \end{subfigure}
     \caption{The  $ppl_{\text{fold}}$, $sr_{\text{fold}}$ and $cov_{\text{fold}}$ performances of different models in three ontologies, over 4 bins of increasing sequence-identity ranges. Low sequence identity indicates low similarity between sequences in a test fold and the training set.  Sequence statistics over the bins (gray dots connected in dashed lines) are also provided. }
     \label{fig:dtm}
\end{figure*}

\section{Comparison of  Folded Structures}  \label{append: structure}
\begin{figure*}[!htb]
    \centering
    \includegraphics[width=1.0\textwidth]{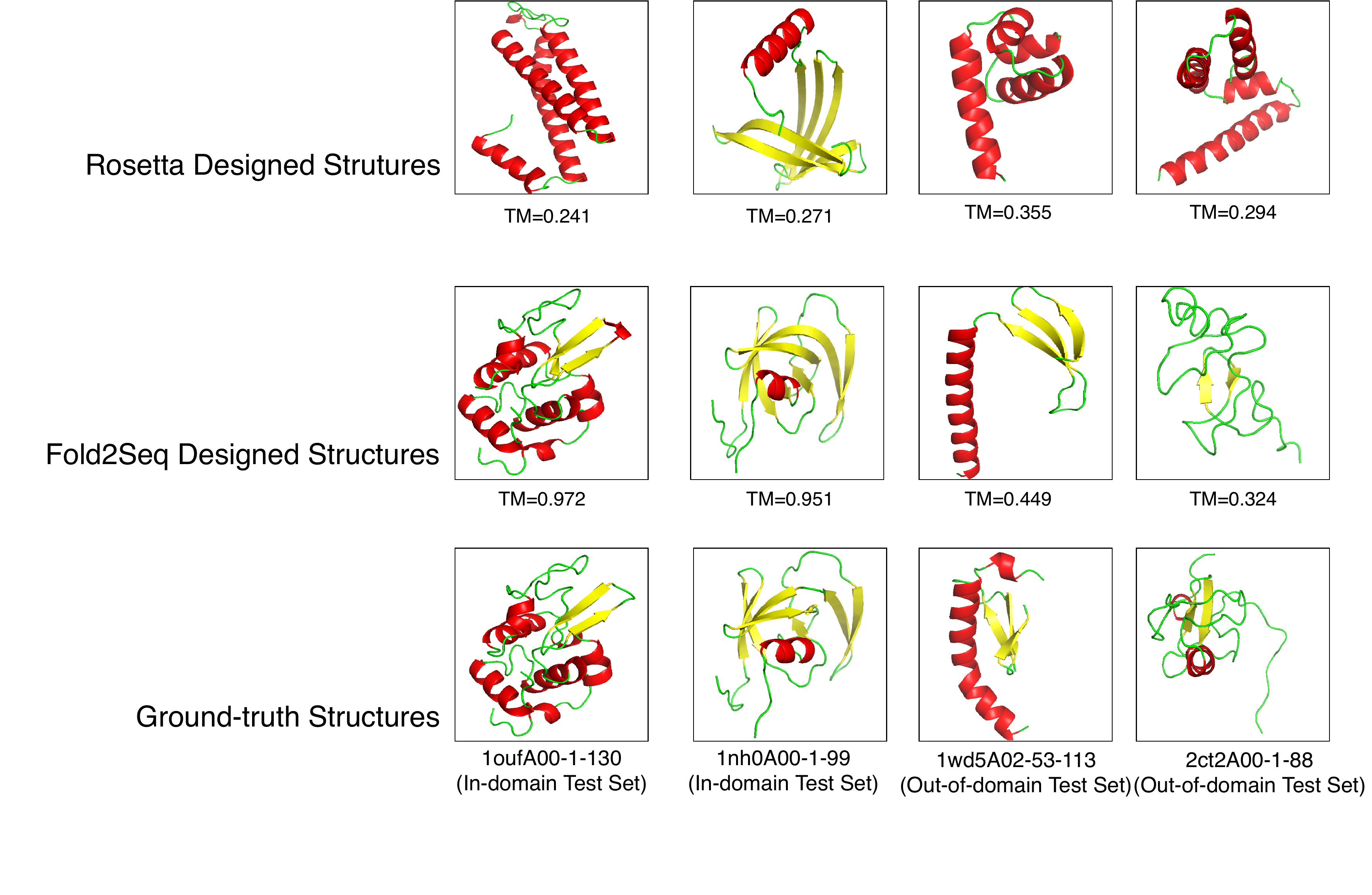}
    \caption{The native and designed structures in the folds with $\Delta tr_{\text{fold}}>0$. The IDs at the bottom are the CATH domain names of each structure. }
    \label{fig:visual1}
\end{figure*}

\begin{figure*}[!htb]
    \centering
    \includegraphics[width=1.0\textwidth]{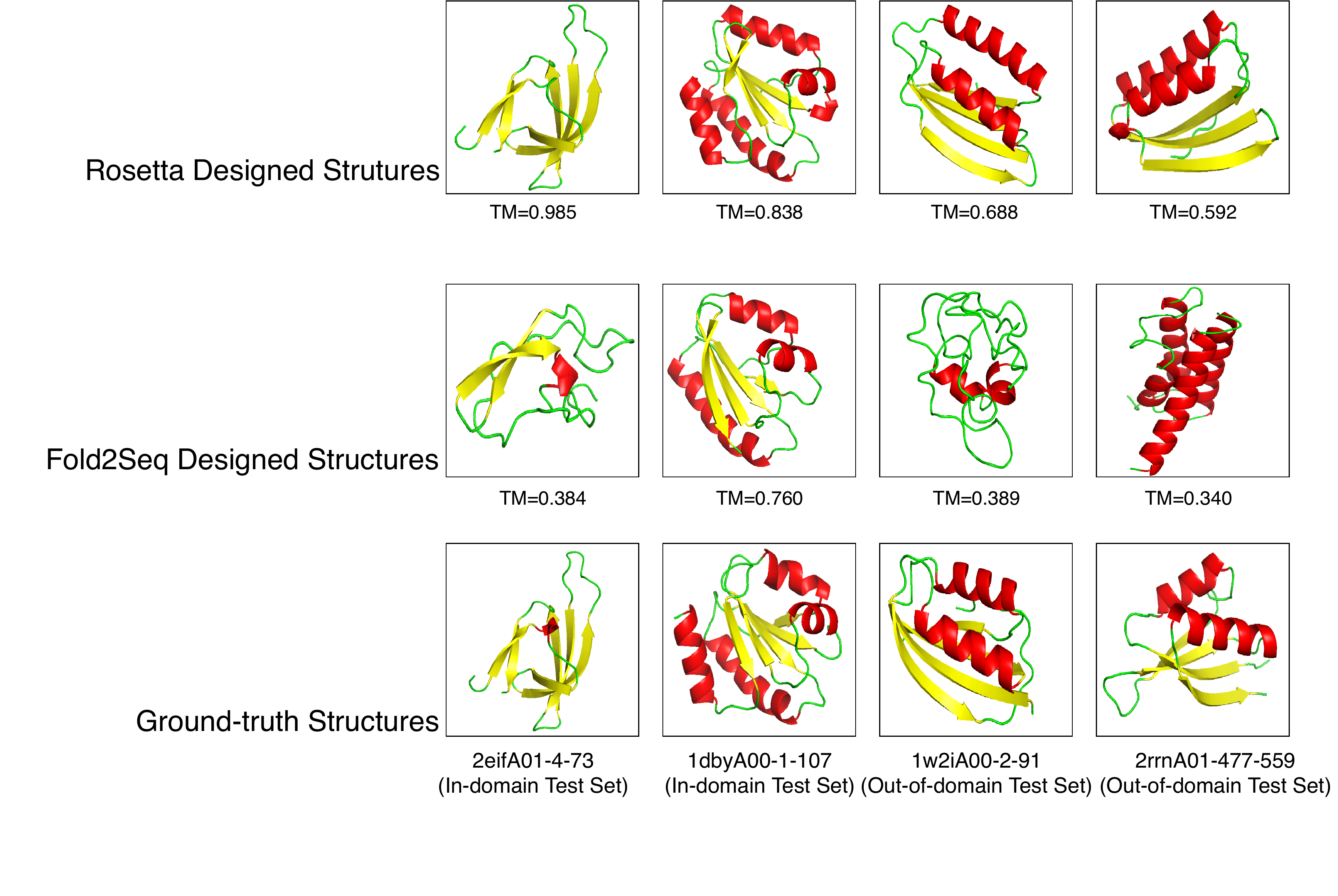}
    \caption{The native and designed structures in the folds  with $\Delta tr_{\text{fold}}<0$. The IDs at the bottom are the CATH domain names of each structure. }
    \label{fig:visual2}
\end{figure*}

In this section, we show some representative folded structures whose sequences are  designed by RosettaDesign and Fold2Seq. The folded structures were predicted using iTasser, a state of the art program for protein structure prediction. Figure~\ref{fig:visual1} shows some structures where Fold2Seq performs better than RosettaDesign and Figure~\ref{fig:visual2} shows some structures where RosettaDesign is better.

\newpage


\newpage
\section{t-SNE Visualization of Fold/Structure Embeddings} 
\label{append: tsne}

\begin{figure*}[!h]
    \centering
    \begin{subfigure}[b]{0.45\textwidth}
     \includegraphics[width=\textwidth]{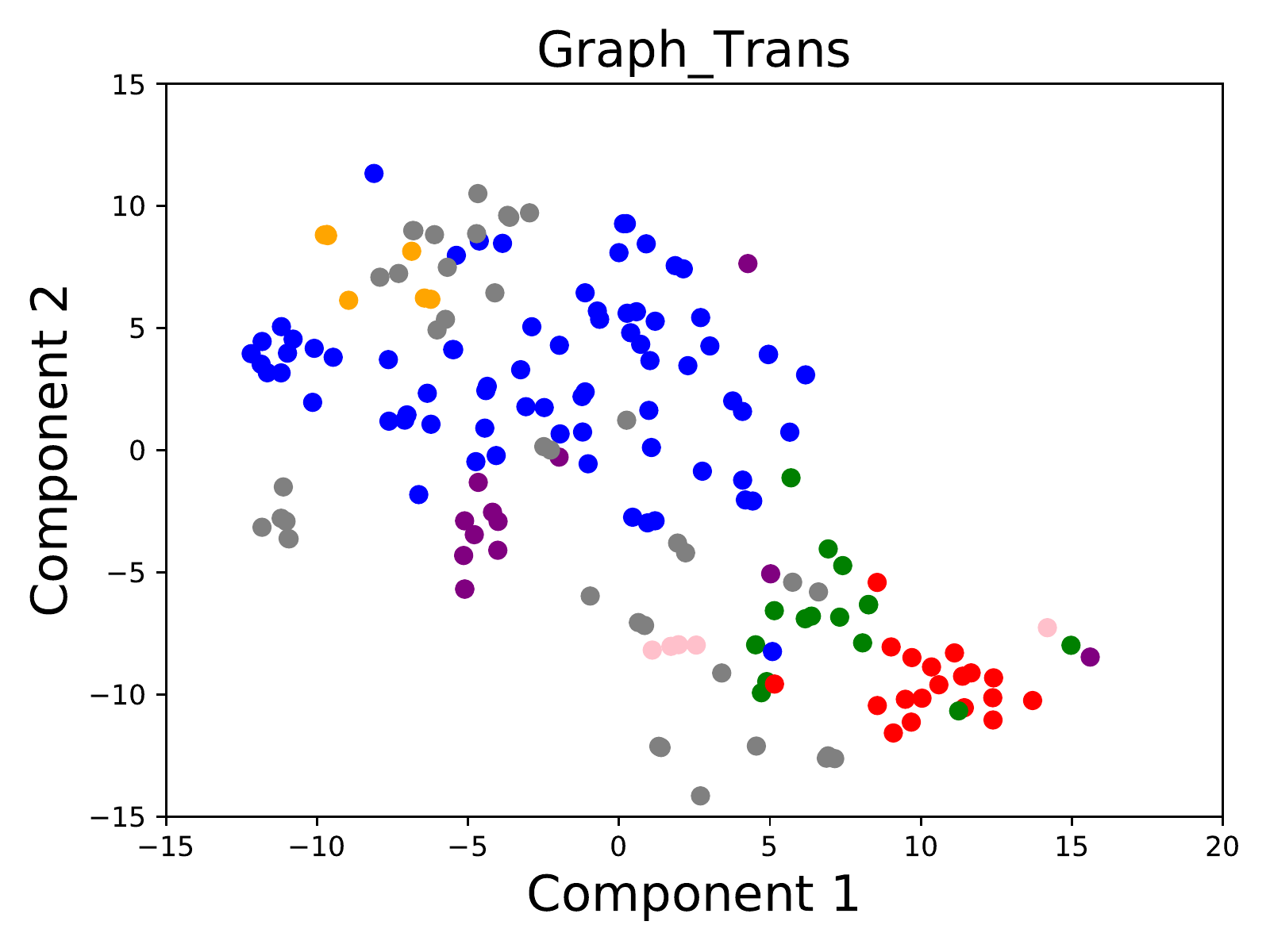}
     \end{subfigure}
    \begin{subfigure}[b]{0.45\textwidth}
    \includegraphics[width=\textwidth]{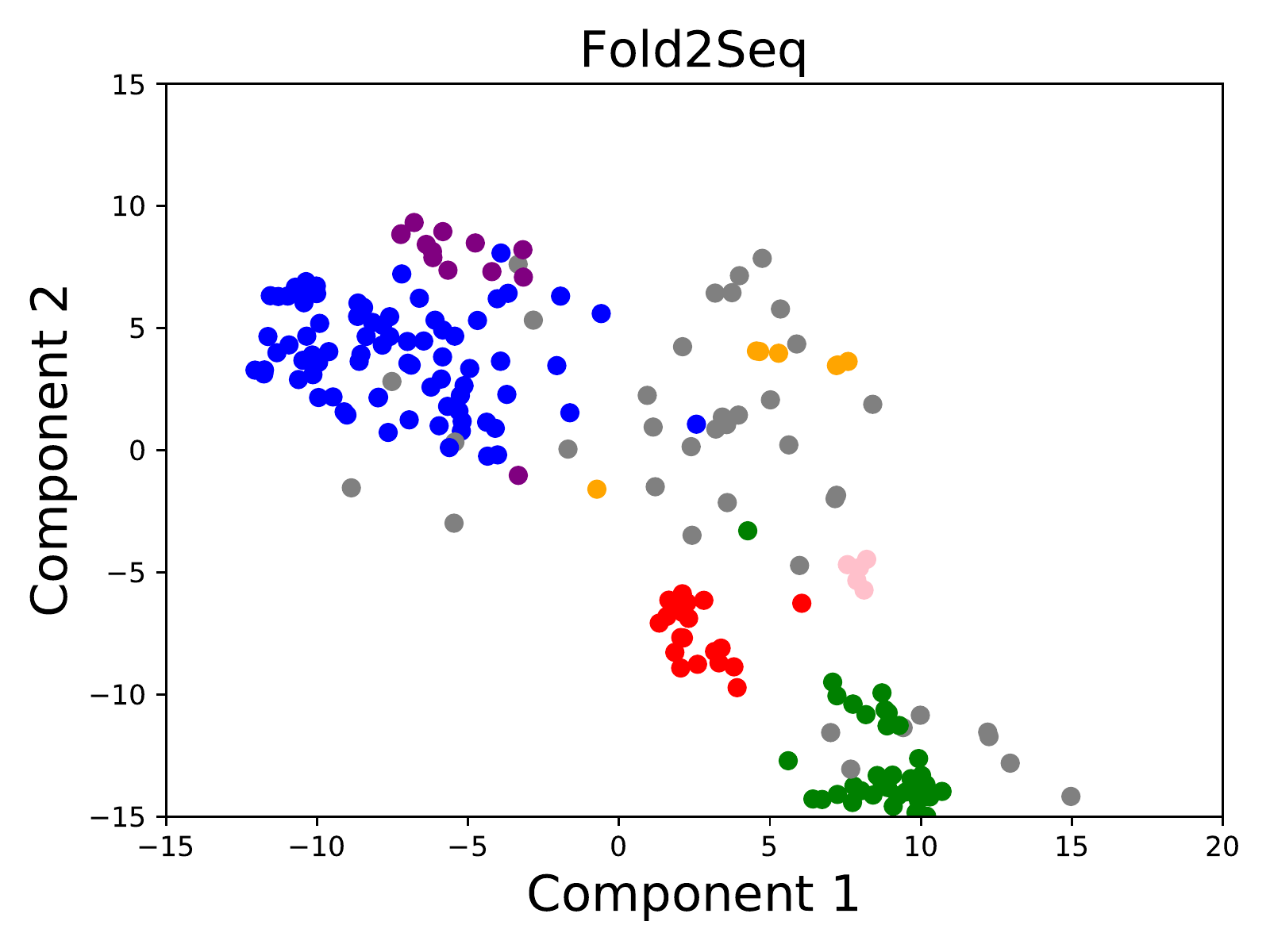}
    \end{subfigure}
    \caption{The t-SNE visualization of the averaged structure (fold) latent embeddings $\bm{h}$ by 
    two methods on the OD test set. Each protein is colored by its fold category. Same color indicates the same fold, except that gray points represent outliers, which is defined by its fold having $<5$ proteins in the test set.
    }
    \label{fig:my_label}
\end{figure*}

We use t-SNE to visualize the fold embeddings $\bm{h}$ from Fold2Seq and Graph\_trans for the proteins in the OD test set (see Figure \ref{fig:my_label}). We show that Fold2Seq captures the similarity and diversity within the fold space better and that the embeddings from proteins belonging to the same fold are better clustered in Fold2Seq.  
 
 \newpage
 \section{Ablation Study} \label{app:abl}

 We performed an ablation study to delineate the contributions from different components of the algorithm. The details of the different ablations are given below.
 
 \begin{itemize}
\item cVAE: We use cVAE \citep{greener2018design} as baseline with 1D string fold representation and MLP-based VAE.
\item Trans\_string\_$\textbf{RE}_f$: We replace the MLP-based VAE in cVAE with transformer autoencoder model. The loss is $L=\textbf{RE}_f$.
\item Trans\_voxel\_$\textbf{RE}_f$: We replace the 1D string fold representation in ``Trans\_string\_$\textbf{RE}_f$" with 3D voxel representation. We also add the convolutional residual  block and 3D positional encoding. The loss is  $L=\textbf{RE}_f$.
\item  +$\textbf{RE}_s$+$\textbf{CS}$:  
We  add the sequence encoder, together with the reconstruction loss and the cosine similarity loss to the previous loss: $L=\lambda_1\textbf{RE}_f + \lambda_2 \textbf{RE}_s  - \lambda_5\textbf{CS}$.
\item +2\textbf{FC}: We add the two \textbf{FC} losses. $ L = \lambda_1 \textbf{RE}_f + \lambda_2 \textbf{RE}_s + \lambda_3 \textbf{FC}_f + \lambda_4 \textbf{FC}_s - \lambda_5\textbf{CS} $.
\item
+\textbf{CY} (Fold2Seq): We add the cyclic loss into the former model with the final loss $ L = \lambda_1 \textbf{RE}_f + \lambda_2 \textbf{RE}_s + \lambda_3 \textbf{FC}_f + \lambda_4 \textbf{FC}_s + \lambda_5 (\textbf{CY} - \textbf{CS} )$. 
\end{itemize}

The key results of the ablation study are summarized in Table 3(a) and Section 4. Overall, the string to voxel change, the addition of 2 \textbf{FC} losses and the cyclic loss gives us a significant performance boost. 





\end{appendices}

\end{document}